\documentclass[11pt]{article}

\usepackage[preprint]{acl}

\usepackage{times}
\usepackage{latexsym}
\usepackage[T1]{fontenc}
\usepackage[utf8]{inputenc}
\usepackage{microtype}
\usepackage{inconsolata}
\usepackage{graphicx}
\usepackage{booktabs}
\usepackage{xspace}
\usepackage{multirow}
\usepackage{tabularx}
\usepackage{makecell}
\usepackage{amsmath}
\usepackage{amssymb}
\usepackage{url}
\usepackage{hyperref}
\usepackage{bbding}
\usepackage[dvipsnames]{xcolor}
\usepackage{arydshln}
\usepackage{fontawesome5}
\usepackage{subcaption}
\usepackage{circledsteps}
\usepackage{placeins}
\usepackage[listings]{tcolorbox}
\usepackage{stfloats}

\tcbuselibrary{breakable, skins}

\newtcolorbox{samplesnippet}[1][]{
    enhanced,
    colback=white,       
    colframe=black,      
    coltitle=black,      
    colbacktitle=white,  
    titlerule=0.5pt,     
    boxrule=0.5pt,       
    toprule=1pt,         
    bottomrule=1pt,      
    leftrule=0pt,        
    rightrule=0pt,       
    arc=0pt,             
    breakable,           
    title=#1             
}

\makeatletter
\newtcblisting[
    list inside=prompt,
    auto counter,
    number within=section
]{prompt}[1][]{
    listing only,
    breakable,
    enhanced,
    colbacktitle=black!60,
    fonttitle=\small,
    coltitle=white,
    before upper=\ttfamily,
    fontupper=\footnotesize,
    left=0pt,
    right=0pt,
    top=0pt,
    bottom=0pt,
    boxsep=4pt,
    boxrule=1pt,
    halign upper=justify,
    fonttitle=\normalfont\small\bfseries,
    title={Prompt~\thetcbcounter},
    title after break={
        \kvtcb@title\space\textnormal{(continued)}
    },
    lines before break=4,
    listing options={
        basicstyle=\ttfamily\footnotesize,
        breaklines=true,
        breakindent=0pt
    },
    #1
}
\makeatother

\newcommand{\bench}{\textsc{CanLegalRAGBench}\xspace}

\newcommand{\metric}{\mathrm{Score}}
\newcolumntype{Y}{>{\raggedright\arraybackslash}X}

\newcolumntype{L}[1]{>{\raggedright\arraybackslash}p{#1}}
\makeatletter
\newcommand{\codestatement}{%
  \ifacl@finalcopy
    Code: \faGithub~\url{https://github.com/NLP-UBC/CanLegalRAGBench}.%
  \else
    The code will be released upon publication.%
  \fi
}
\makeatother

\title{{\color{red}\faIcon{canadian-maple-leaf}}\bench:\\Evaluating Retrieval-Augmented Generation on Canadian Case Law}

\author{Ethan Zhao$^{1,2}$~~~Maksym Taranukhin$^{1,2}$~~~Wei Cui$^4$~~~Moira Aikenhead$^4$~~~Vered Shwartz$^{1,2,3}$\\\\
$^1$ Department of Computer Science, University of British Columbia\qquad
$^2$ Vector Institute\\
$^3$ CIFAR AI Chair\qquad
$^4$ Peter A. Allard School of Law, University of British Columbia\\
\texttt{\small \{ethanz01, maksymt, vshwartz\}@cs.ubc.ca
}}

\begin{document}
\maketitle

\begin{abstract}
    RAG-based legal assistants have been growing in popularity, but LLM hallucinations remain a key issue and potentially undermine justice. While benchmarks have been developed to evaluate progress, many rely on unnatural queries rather than realistic legal scenarios. Moreover, Canadian law remains underrepresented in existing evaluations. To address this gap, we introduce \bench{}, a Canadian legal QA benchmark based on 532 expert-validated realistic style queries and expert-annotated answers grounded in case law. Our evaluation shows that retrieval performance is sensitive to RAG design choices and that open-source embedding models can be competitive with closed-source models. However, it also reveals the limitation of automatic evaluations that penalize systems for retrieving alternative relevant documents. We also find that generated answers often diverge from gold responses, either with hallucinations or in style, often producing overly detailed or irrelevant content, with 8-29\% of claims not being supported by the retrieved documents. We hope this benchmark will help drive continued progress in addressing limitations of legal RAG systems. \codestatement

\end{abstract}

\section{Introduction}
\label{sec:intro}
Retrieval-augmented generation \cite[RAG;][]{rag} is a natural fit for legal question answering, since legal answers should be grounded in authorities that can be inspected and cited. RAG-based legal assistants increasingly help laypeople and self-represented litigants understand their rights and help legal professionals research more efficiently. Yet LLM-based legal assistants hallucinate unsupported claims and readily accept users' incorrect assumptions \cite{hallucination}. Such errors force professionals to spend significant time verifying claims and risk distorting justice.

To measure progress on that front, researchers have built benchmarks for legal language understanding \cite{chalkidis-etal-2022-lexglue,guha2023legalbench}, case retrieval and entailment \cite{rabelo2022coliee}. Recent benchmarks have begun to address retrieval and end-to-end grounded generation \cite{mahari-etal-2024-lepard,zheng2025,hou-etal-2025-clerc} but most benchmarks introduce artificial tasks -- such as reproducing references from citations -- rather than simulating realistic uses such as legal research. COLIEE \cite{coliee2025overview}, the only Canadian case retrieval task we found, evaluates systems on finding supporting prior case laws for a given court case. To the best of our knowledge, there are no public benchmarks for Canadian case law retrieval where queries simulate realistic legal research queries. To bridge this gap, we propose \bench{}, a benchmark with synthetic queries simulating legal questions from users with different levels of legal expertise, and expert-annotated answers grounded in Canadian case law. 

We evaluated a range of RAG methods on our benchmark. We show that dense retrieval and enhancements such as reranking outperform sparse indexing, chunk size matters, and open-source embedding models are competitive with close-source models. However, we also reveal a limitation of standard automatic retrieval evaluations in the legal setting, where many case laws may be relevant for a given legal query, but even a subset may be enough for writing a sound answer. We show that these evaluations penalize systems for retrieving alternative relevant documents, and argue for the necessity of expert evaluation in this domain.

\begin{table*}[t]
    \centering
    \footnotesize
    \setlength{\tabcolsep}{2pt} 
    \begin{tabular}{llllllrr}
        \toprule
        \textbf{Benchmark}    & \textbf{Jurisdiction} & \textbf{Lang} & \textbf{Corpus} & \textbf{Retrieval Task} & \textbf{Supervision}  & \textbf{\# Docs} & \textbf{\# Queries} \\
        \midrule
        \textbf{BSARD} & Belgium & FR & Law articles    & Common Questions & Lawyer & 22.6k 
        & 1,108 \\
        \textbf{EQUALS} & China & ZH & Law articles & Online Questions & Law student & 3,081 & 6,914 \\
        \textbf{LegalBench-RAG} & US & EN & Documents & Questions about Docs & Automated & 714 & 6,858 \\
        \midrule
        \textbf{LePaRD} & US & EN & Case law    & Context to Quotation & Automated & 1M 
        & 4.3M \\
        \textbf{CLERC} & US & EN & Case law    & Case block to Citation & Automated & 1.8M 
        & 105k       \\
        \textbf{COLIEE} & Canada & EN & Case law    & Case to Citation & Automated &  1,470  
        & 400        \\
        \midrule
        \textbf{\bench{}}     & Canada & EN & Case law    & Legal Research & Law student & 588  & 532 \\
        \bottomrule
    \end{tabular}
    \vspace{-5pt}
    \caption{    Summary of existing legal retrieval benchmarks.
    BSARD~\cite{louis-spanakis-2022-statutory} answers commonly asked
    questions submitted to a nonprofit legal service.
    EQUALS~\cite{Chen-et-al-2023-EQUALS} answers legal questions from
    online forums.
    LegalBench-RAG~\cite{pipitone2024legalbenchrag} answers questions
    about documents.
    LePaRD~\cite{mahari-etal-2024-lepard} takes the context preceding a
    quotation in a court decision and retrieves the quoted document.
    CLERC~\cite{hou-etal-2025-clerc} takes text blocks from court decisions
    and retrieves cited documents.
    COLIEE~\cite{coliee2025overview} takes full court decisions and
    retrieves cited documents.
    \bench{} answers synthetic legal-research questions grounded in facts
    from court decisions.}
    \label{tab:benchmark-comparison}
\end{table*}

Generated answers differed substantially from the gold standard answers, especially when these answers were based on a different subset of relevant case laws. Manual analysis revealed that models generated long answers that often contained irrelevant, overly-detailed claims that miss the point of the query. Finally, at least 20\% of the generated claims were unsupported by the documents they were referencing, 40\% of which were due to hallucination and misapplication of the law. 
We hope this benchmark will support future work in addressing remaining gaps in legal RAG systems.

\section{Background}
\label{sec:background}
Many existing legal benchmarks focus on narrow legal contexts -- such as US merger agreements \cite{wang-etal-2023-maud}; but there are also evaluation suites such as LexGLUE \cite{chalkidis-etal-2022-lexglue}, LegalBench \cite{guha2023legalbench}, and LawBench \cite{fei-etal-2024-lawbench} that test various legal contexts and capabilities such as legal classification, extraction, NLI, and QA. 
Classification tasks most commonly involve legal judgment prediction \cite{niklaus-etal-2021-swiss,masala-etal-2021-jurbert,semo-etal-2022-classactionprediction}. Information extraction tasks include named entity recognition and terminology extraction \cite{pais-etal-2021-named,pham-etal-2021-legal,kalamkar-etal-2022-named}, and extraction of structured information from documents \cite{simonson-2021-supervised,wenger-etal-2021-automated,cuad}. As models' reasoning abilities improved, benchmarks were developed to measure legal reasoning, for instance identifying legal violations 
\cite{hagag-etal-2024-legallens}. 

More recent work has shifted toward legal question answering. BarExamQA \cite{zheng2025} consists of multiple-choice US bar exam questions with supporting gold passages from case laws, online resources, and textbooks. 
Of particular relevance to our work are benchmarks that evaluate retrieval and answer generation. We situate our benchmark against those in Table~\ref{tab:benchmark-comparison}. 
A few benchmarks focus on legal case retrieval \cite{mahari-etal-2024-lepard, hou-etal-2025-clerc, coliee2025overview}, which is a crucial component of legal reasoning in common law jurisdictions, where outcomes rely on precedent. 

What these benchmarks have in common is that they use the structure of existing legal documents to construct a large-scale automatically labeled benchmark -- for example, predicting a relevant case cited in the original case. These tasks are less representative of actual use of legal research tools. Our benchmark addresses this gap by combining natural language queries \cite{ravichander-etal-2019-question, louis-spanakis-2022-statutory, Chen-et-al-2023-EQUALS} with Canadian case law retrieval documents, enabling more realistic evaluation of legal retrieval-augmented generation.

\begin{figure*}[t]
    \centering
    \includegraphics[width=0.8\textwidth]{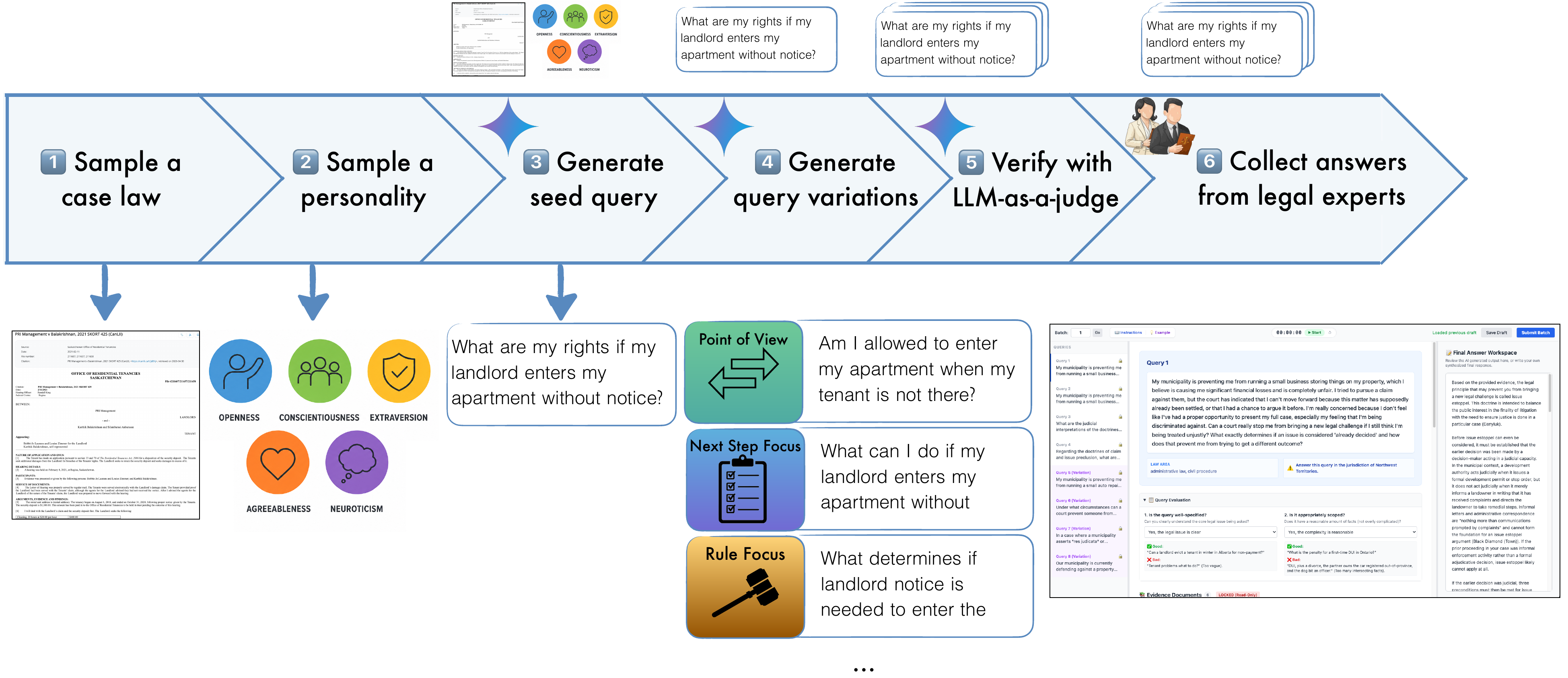}
    \caption{The pipeline used for creating \bench{}.}
    \vspace{-5pt}
    \label{fig:data_collection_process}
\end{figure*}

\section{\bench}
\label{sec:data}

We introduce \bench{}, a benchmark comprising of legal queries grounded in Canadian case law. The goal of \bench{} is to evaluate RAG-based systems on natural language queries based on facts and scenarios from real cases, simulating a common use of legal AI assistants. Access to real legal queries is unavailable for privacy reasons, so we instead use expert-validated synthetic queries. To create the dataset, we employed a 6-step process, as illustrated in Fig.~\ref{fig:data_collection_process} and detailed below.

\paragraph{\Circled{1} Sample a case law.} Each seed query in our benchmark is based on an existing case. We use the A2AJ Open Canadian Legal Data \cite[CLD;][]{a2aj}, a publicly available collection of Canadian court decisions. To remove noisy documents and keep the computational requirements reasonable, we filtered out documents in the top or bottom quartile in terms of length, keeping only the middle 50\%.
We selected a subset of the courts and tribunals that represent common legal disputes encountered by the general public, and sampled documents equally.  

\paragraph{\Circled{2} Sample a user persona.} User personas were sampled using low, medium, and high amounts of each personality from the Big Five personality traits  \cite{5_personality}. Different prompts given to the query generator were used to vary this. For all query generations we use DSPy \cite{khattab2023dspycompilingdeclarativelanguage} to template our prompts.

\paragraph{\Circled{3} Generate seed query.} Given a case law and a user persona, we prompt \texttt{Gemini 2.5 Flash} to simulate a legal query from this user that can be answered by the given case law. See Prompt~\ref{prompt:query_generation} in the appendix. To get provincial coverage, we sampled from all provinces and territories except Quebec which follows civil law (see Appendix~\ref{appendix:additional_info:provinces} for the composition). To add provinces to queries, we simply prepend ``I am in \{province\}.''

\paragraph{\Circled{4} Generate query variations.} We prompted \texttt{Gemini 2.5 Flash} to generate 8 variations of each seed query. We generate four variations where the facts remain the same -- thus a robust system is expected to retrieve the same case laws -- but where the focus of the question or the user persona change. For example, the seed query ``What are my rights if my landlord enters my apartment without notice?'' in Fig.~\ref{fig:data_collection_process} can also be asked from the point of view of a landlord checking if they are allowed to enter their apartment when the tenant is not there. The other four variations change a small fact in the seed query, which may require systems to retrieve different case laws. Finally, within each batch of 8 queries, half of the queries simulate a legal professional and half simulate a lay person. See the appendix for prompts  \ref{prompt:query_variation_alternative_facts} and \ref{prompt:query_variation_alternative_situation}, and query variations (Appendix~\ref{appendix:additional_info:variations}). 

\paragraph{\Circled{5} Verify with LLM-as-a-judge.} In order to reduce the cost of manual annotation, we used \texttt{Gemini 2.5 Pro} to filter out $\sim$20\% of queries that are self-contradictory, nonsensical, overly complex, hinting towards the document with references, or otherwise unfit for a retrieval setup. See Prompts~\ref{prompt:judge-system} and \ref{prompt:judge-user} in the appendix.  

\paragraph{\Circled{6} Collect answers from legal experts.} We hired law students from Canadian universities 
and a Canadian paralegal from Upwork to provide the ground truth answers for our benchmark. Annotators took 2-3 hours to complete a batch of 8 queries, and were paid \$20 CAD per hour. To ensure the quality of annotations, 
we manually reviewed and filtered around 30 examples with obvious errors, such as answers that fail to target the query. We report full data statistics in Appendix~\ref{appendix:full_dataset_stats}.

\section{Retrieval}
\label{sec:retrieval}
We present the document collection (\S\ref{sec:retrieval:doc_collection}), the baselines (\S\ref{sec:retrieval:baselines}), evaluation metrics (\S\ref{sec:retrieval:eval_metrics}), and results (\S\ref{sec:retrieval:results}) for the retrieval task, and analyze the performance and errors  (\S\ref{sec:retrieval:analysis}).

\subsection{Document Collection and Chunking} 
\label{sec:retrieval:doc_collection}

To facilitate testing RAG systems on our benchmark, \bench{} includes additional distractor case laws, randomly selected in equal proportions from CLD and a second dataset, which have been preprocessed and filtered as described in Sec.~\ref{sec:data}. Since at the time of construction, CLD only comprised 11 sources (generally covering BC, Ontario, federal, and refugee law), we supplemented it with a second dataset of public court decisions across Canadian jurisdictions, which was provided to the authors by a private company.\footnote{The portion of the data used in this work will be made available upon publication.} 
This dataset covers additional jurisdictions and more diverse legal issues. We applied various strategies to remove duplicate case laws, including citation normalization, case normalization, stripping whitespaces, and manually comparing documents from the same year and similar lengths. The dataset details are provided in Table~\ref{tab:corpus-stats}. 
All chunking is implemented as follows. We test various values of max character sizes (1024--8192). We chunk documents via recursive text splitting with 128 character overlap. The text splitter uses hierarchical text breaks such as newlines, periods and spaces.


\subsection{Baselines and Evaluated Models}
\label{sec:retrieval:baselines}

We benchmark various retrieval systems on \bench{}, including sparse retrieval, dense retrieval, and enhancements such as reranking and IterRetGen, which are detailed below. Table~\ref{tab:models} lists all the models we evaluated. 

\paragraph{Sparse Retrieval.} We use BM25, an established keyword-based retrieval algorithm which is still considered a strong baseline \cite{bm25}. Since legal documents are generally long, we use the BM25L variant, which is specifically designed for long documents \cite{bm25L}. 

\paragraph{Dense Retrieval.} Dense retrieval works by embedding document chunks and retrieving based on vector similarity \cite{rag}. Our choice of embedding models is based on high ranking on the legal embedding benchmark \cite{mleb}, and includes strong general-purpose, open-weights LLMs \texttt{Qwen 3 Embedding 8B} \cite{zhang2025qwen3embeddingadvancingtext} and \texttt{EmbeddingGemma} \cite{vera2025embeddinggemmapowerfullightweighttext} and proprietary LLMs \texttt{Gemini Embedding 001, Gemini Embedding 2} \cite{lee2025geminiembeddinggeneralizableembeddings, shanbhogue2026geminiembedding2native} and a legal-specific \texttt{Kanon 2} \cite{mleb}. 

\paragraph{Reranking.} Reranking is a common way to improve retrieval methods. A typical reranker takes the query and retrieved documents and processes them together using a cross-encoder transformer to output a relevance score for each pair. We use the proprietary \texttt{Kanon 2 Reranker} and implement reranking by retrieving $4\times k$ documents, reranking, and retaining the top k for evaluation.

\paragraph{Hybrid Retrieval.} We retrieve the top k documents for each of the sparse and dense retrievers, deduplicate them, rank using reciprocal rank fusion \cite{Cormack-RecipricalRankFusion-2009} or the  \texttt{Kanon-2}  reranker, and retain the top k documents. 

\paragraph{Iterative Retrieval.} Iter-RetGen \cite[IRG;][]{shao-etal-2023-enhancing} is a retrieval enhancement that iteratively generates an answer and uses it in the retrieval query, motivated by the intuition that an answer typically has higher overlap with the documents than the query. We run the method for 3 iterations and use \texttt{Gemini Flash-2.5} for answer generation (see Prompt~\ref{prompt:iter-retgen} in the appendix).

\newcolumntype{R}{>{\raggedleft\arraybackslash}X}

\begin{table*}[t]
\centering
\footnotesize
\begin{tabular*}{\textwidth}{@{\extracolsep{\fill}} lrlcrrrrr}
\toprule
\textbf{Model} & \textbf{Chunk} & \textbf{Method} & \textbf{Rerank} & \textbf{MRR} & \textbf{Recall@10} & \textbf{Recall@25} & \textbf{nDCG@10} & \textbf{nDCG@25} \\
\midrule
-- & Doc & Sparse (BM25) & \textcolor{RedOrange}{\XSolidBrush} & 0.245 & 0.193 & 0.309 & 0.163 & 0.208 \\
Qwen & 1024 & Hybrid & \textcolor{ForestGreen}{\Checkmark} & 0.563 & 0.364 & 0.548 & 0.426 & 0.477 \\
\midrule
Qwen & 8192 & IterRetGen & \textcolor{ForestGreen}{\Checkmark} & \underline{0.632} & 0.406 & 0.601 & \underline{0.474} & \underline{0.527} \\
Gemma & 8192 & IterRetGen & \textcolor{ForestGreen}{\Checkmark} & 0.613 & \underline{0.431} & 0.595 & 0.471 & 0.510 \\
Gemini2 & 4096 & IterRetGen & \textcolor{RedOrange}{\XSolidBrush} & \textbf{0.661} & \textbf{0.456} & \textbf{0.631} & \textbf{0.521} & \textbf{0.560} \\
Kanon2 & 8192 & Dense & \textcolor{ForestGreen}{\Checkmark} & 0.590 & 0.406 & \underline{0.624} & 0.449 & 0.514 \\

\bottomrule
\end{tabular*}
\vspace{-5pt}
\caption{Best performing configurations for each embedding model family on \bench.  Full retrieval results are available in \autoref{tab:retrieval_results} in the appendix. \textbf{Bold} = overall best, \underline{underlined} = 2nd best}
\label{tab:main-results}
\end{table*}


\subsection{Evaluation Metrics} 
\label{sec:retrieval:eval_metrics}

\paragraph{Automatic Evaluation.} We report recall@k, MRR \cite[MRR;][]{radev-etal-2002-evaluating}, and nDCG \cite[nDCG;][]{jarvelin2002cumulated}. These metrics measure the ability to find correct documents and rank them appropriately. Following prior work \cite{louis-spanakis-2022-statutory, hou-etal-2025-clerc}, we consider recall and nDCG as the primary metrics (nDCG is more nuanced than MRR as it looks at the ranking of all top-k retrieved documents; and recall is complementary to all other metrics). We report all metrics for the main experiments, and perform analysis based on nDCG and recall, with $k=10$. All aggregated metrics are macro averages, calculated as follows: 
$\metric = \frac{1}{|Q|}\sum_{q \in Q} M(q, D_r^q, D_{GT}^q)$,  
where $Q$ is the set of queries in \bench{}, $M$ is the metric, and $D_r^q$ and $D_{GT}^q$ are the retrieved and gold standard documents for each query. We map retrieved chunks to their documents and report metrics at the document level. 

\paragraph{Expert Evaluation.} Automatic evaluation provides a general comparison of baselines, and has been employed by prior work. We diverge from prior work and identify a mismatch between the \emph{completeness assumption} that automatic metrics make -- i.e., that a correct answer must be based on \emph{all} the relevant documents -- and the nature of legal research. In reality, some queries have a large number of relevant cases, and even lawyers would not read all of them to come up with an answer.\footnote{The exception is questions that ask about general trends (``what is the percentage of cases <description> with <outcome>?'', but our queries are designed to refer to a specific legal issue.} Automatic metrics may penalize systems for retrieving relevant documents that are not in the gold set whereas we would like to reward models for any answer that can be supported by relevant case laws. To that end, diverging from prior work, we also perform expert evaluation. We created a stratified sample of 30 queries by province first and by legal issue second (using \texttt{Gemini-2.5-Pro} for categorization). Expert annotators (\S\ref{sec:data}) evaluated the retrieved documents of 4 best performing configurations according to the automatic evaluation (\texttt{Qwen 8192 chunk size IRG + Rerank}, \texttt{Gemma Embedding 2 8192 chunk size IRG Rerank}, \texttt{Gemini-Embedding-2 4096 chunk size IRG}, and \texttt{Kanon2 8192 chunk size + rerank}).\footnote{We treat the automatic metrics as an imperfect proxy;  better retrieval systems should still on average yield higher document overlap with the gold documents compared to weak retrieval systems.} 
We instructed the expert annotators to judge retrieved documents for relevance to the query. Documents judged as relevant were added to the gold standard documents, and we report the automatic metrics before and after expert evaluation for this subset.

\subsection{Results} 
\label{sec:retrieval:results}

We report the overall retrieval performance on \bench{}, and look into the effect of retrieval method and embedding model on the performance. 

\paragraph{General Trends.} We tested 46 combinations of method, embedding model, and chunk size. The complete results of all evaluated configurations are reported in \autoref{tab:retrieval_results} in Appendix~\ref{appendix:full_results}. Here, we report the best-performing configuration for each embedding model family in \autoref{tab:main-results}, based on the average across evaluation metrics. 

The best dense retrieval-based methods dramatically outperform sparse retrieval, and the hybrid approach seems to also fall behind dense retrieval. Enhancements such as IterRetGen and reranking improve upon a vanilla dense retrieval approach but inconsistently. \textbf{Large chunk sizes generally perform better} -- suggesting a good trade off of preserving legal context while adding noise -- but it is also model and method-dependent. Finally, with respect to model family, \textbf{\texttt{Gemini-2} performs best across most evaluation metrics}, with \texttt{Gemma} and \texttt{Qwen} closely following, suggesting that open-source embedding models are close in quality to closed-source models. Interestingly, the \texttt{Kanon-2} embeddings, which are specialized for the legal domain, slightly underperform other embedding models. 

\begin{figure}
    \centering
    \includegraphics[width=\linewidth]{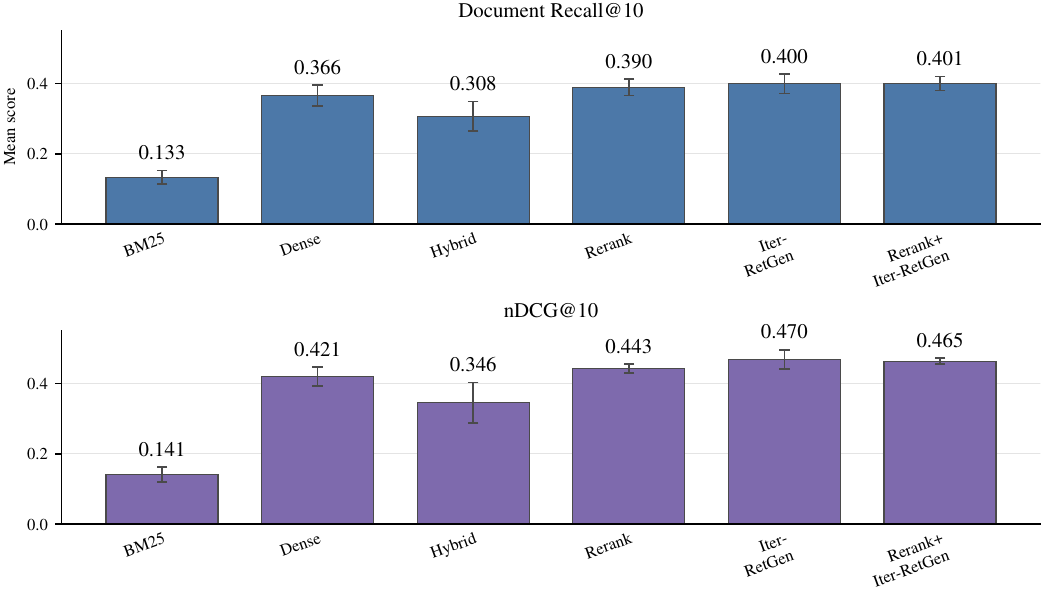}
    \vspace{-15pt}
    \caption{Average recall@10 and nDCG@10 for each retrieval method, aggregated across models and chunk sizes, with standard deviation error bars.}
    \label{fig:retrieval_methods}
\end{figure}

\begin{figure*}
    \centering
    \includegraphics[width=0.8\linewidth]{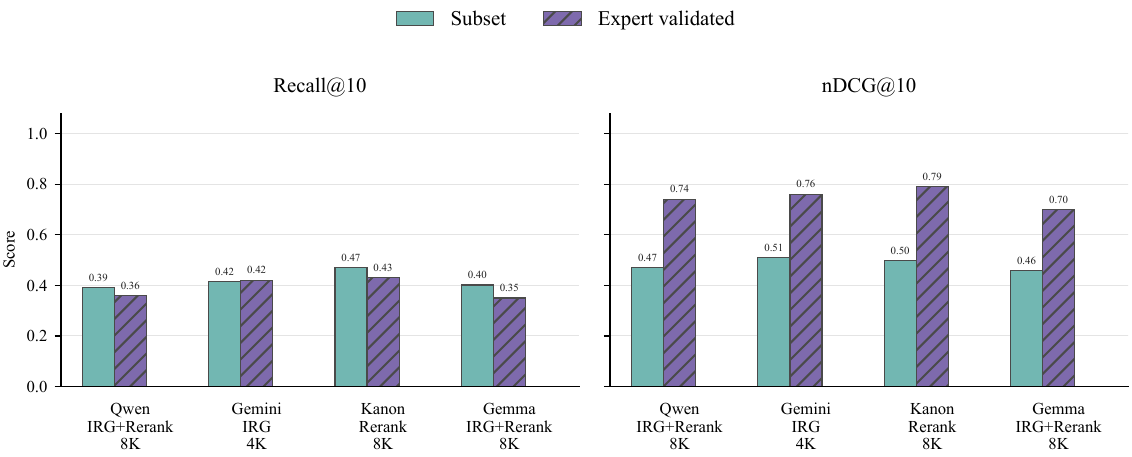}
    \caption{Retrieval metrics before and after expert evaluation of document relevance on a subset of 30 queries.}
    \vspace{-5pt}
    \label{fig:human_eval}
\end{figure*}

\paragraph{Best retrieval method.} We report the performance of each retrieval method averaged across embedding model and chunk size in Fig.~\ref{fig:retrieval_methods}. Reranking, IterRetGen, and their combination, perform highly. Vanilla dense retrieval is marginally behind. Sparse and hybrid retrieval underperform. The hybrid method performs worse than dense retrieval on both metrics due to poor BM25 performance.

Table ~\ref{tab:retrieval_results} shows that the effect of reranking on performance is configuration-dependent. It consistently improves \texttt{Kanon-2} across all chunk sizes and metrics --  this is unsurprising given that the reranker is based on \texttt{Kanon}, so they are presumably designed to work well together. However, it is mostly useful for \texttt{Qwen} and \texttt{Gemma} with larger chunks -- potentially because they can be noisier and reranking can help filter the noise. 
IRG provides the highest gains when the base model is already strong (e.g., the \texttt{Gemini} family). IRG by itself performs quite strongly
(Fig.~\ref{fig:retrieval_methods}). 

\paragraph{Expert evaluation.} Fig.~\ref{fig:human_eval} displays the recall@10 and nDCG@10 of the best configurations before and after expert evaluation of document relevance on a subset of 30 queries. Recall surprisingly decreased for most methods; 
Conversely, nDCG improves substantially, since high-ranking documents that were considered irrelevant may now be contributing to the DCG (the numerator of nDCG). This shows that non-trivial amounts of penalized retrieved documents were indeed relevant. The distribution happens to reduce recall for most systems.
With that said, the evaluation suggests that the automatic and expert evaluations predict a similar trend, maintaining relative rankings on all except 1 pair for each metric. And the pairs that were swapped both had a small 1-point start difference.


\textbf{Overall, these results suggest that automatic metrics predict a similar trend as expert evaluation, but underestimate the absolute performance of systems.} 


\begin{table}[t]
\centering
\footnotesize
\begin{tabularx}{\linewidth}{@{} >{\raggedright\arraybackslash}X @{}}
\toprule
\textbf{Query:} \ttfamily Ex won't leave my house, keeps coming back in after I tell them no. Even after police! What happens to people who keep breaking into your home like that? Is there a real penalty? How do I stop it? It's my house! \\
\midrule
\textbf{Retrieved Chunk 1 (Qwen \& Kanon 2):} {[omitted details...]} \texttt{The police, during a murder investigation in 1991, entered the accused's house (an equipment trailer) without permission. When they received no answer at the door, they entered, roused the accused, touched his leg, ordered him to get up and took him to the front of the trailer for better lighting. The police arrested him after seeing blood on his shirt.} {[omitted details...]} \\
\midrule
\textbf{Retrieved Chunk 2 (Gemini-2):} {[Omitted Details...]} \texttt{3) In the stairwell of Q's apartment the accused} {[Omitted graphic detail...]} {and threatened to kill her} {[Omitted Details...]} \texttt{The offences represent a sustained pattern of physical and mental abuse of an intimate partner over a four‑year period, and an effort to avoid responsibility through efforts to obstruct justice.} \\
\bottomrule
\end{tabularx}
\vspace{-5pt}
\caption{Retrieved irrelevant chunks for example query.}
\label{tab:error_example}
\end{table}
\begin{table}[t]
\centering
\footnotesize
\setlength{\tabcolsep}{2pt} 
\begin{tabular}{llcccr}
\toprule
\textbf{Config.} & \textbf{Subset} & \textbf{R@10} & \textbf{nDCG@10} &  \textbf{Ground} & \textbf{Acc} \\
\midrule

\multirow{6}{*}{Gemini} 
 & Layperson    & 0.461 & 0.505 & 0.747 & 0.422 \\
 & Expert       & 0.451 & 0.537 & 0.784 & 0.444 \\
\cmidrule{2-6}
 & Ontario      & 0.475 & 0.526 & 0.724 & 0.447 \\
 & BC           & 0.445 & 0.494 & 0.761 & 0.453 \\
 & Alberta      & 0.428 & 0.519 & 0.755 & 0.435 \\
 & Other        & 0.467 & 0.538 & 0.822 & 0.401 \\  
\midrule

\multirow{6}{*}{Gemma} 
 & Layperson    & 0.443 & 0.468 & 0.767 & 0.478 \\
 & Expert       & 0.418 & 0.475 & 0.723 & 0.504 \\
\cmidrule{2-6}
 & Ontario      & 0.458 & 0.497 & 0.764 & 0.503 \\
 & BC           & 0.351 & 0.429 & 0.696 & 0.536 \\
 & Alberta      & 0.440 & 0.508 & 0.771 & 0.477 \\
 & Other        & 0.459 & 0.451 & 0.748 & 0.461 \\ 
\midrule
 
\multirow{6}{*}{Qwen} 
 & Layperson    & 0.389 & 0.442 & 0.704 & 0.435 \\
 & Expert       & 0.423 & 0.508 & 0.717 & 0.414 \\
\cmidrule{2-6}
 & Ontario      & 0.438 & 0.514 & 0.743 & 0.461 \\
 & BC           & 0.387 & 0.463 & 0.701 & 0.424 \\
 & Alberta      & 0.386 & 0.492 & 0.716 & 0.445 \\
 & Other        & 0.404 & 0.431 & 0.681 & 0.356 \\
\bottomrule

\end{tabular}
\caption{Performance per query type and jurisdiction for the best-performing configuration overall for each embedding family.}
\vspace{-5pt}
\label{tab:performance_analysis}
\end{table}

\subsection{Analysis}
\label{sec:retrieval:analysis}

\paragraph{Case Analysis.} Table~\ref{tab:error_example} demonstrates a retrieval example that highlights various error cases such as retrieving documents that are topically similar to the query but differ in legal issue, party roles, and procedural posture. The query asks what happens when an ex-partner repeatedly enters your home despite being unwelcome, even after police involvement. The Qwen and Kanon 2 configurations retrieve the first chunk, which discusses the police entering an accused person's home during a murder investigation, reversing the roles (police entering the home \emph{of the accused}). The second retrieved passage concerns intimate partner violence and threats in an apartment building without discussing entrance. This example motivates better retrieval methods that measure not only topical similarity, but also specific legal logic needed to answer the query. 

\paragraph{Analysis of performance per query type.} Table~\ref{tab:performance_analysis} presents the performance based on query type (legal expert or lay person) and province. Across methods, retrieval performance is generally worse on BC and Alberta and best on Ontario, but  \texttt{Gemini} has the least variation between provinces showing robustness to legal domains. 
Comparing the query types, the nDCG is higher for legal experts across models but recall is mixed. 
We compare the difference in query style performance deltas between base and enhanced configurations in Fig.~\ref{fig:style_recall} and Fig.~\ref{fig:style_ndcg} (Appendix~\ref{appendix:full_results}). In base configurations, performance is always better on queries simulating legal experts, with large gaps in nDCG@10. This illustrates how embeddings are not robust to the difference between layperson language and legal documents. Enhancements bridge the gaps in nDCG@10, likely due to IRG writing intermediate answers using legal terminology.

\section{Generation}
\label{sec:generation}
We present the baselines (\S\ref{sec:generation:baselines}), evaluation metrics (\S\ref{sec:generation:eval_metrics}), and results (\S\ref{sec:generation:results}) for the generation task, and analyze the performance (\S\ref{sec:generation:analysis}).  

\subsection{Baselines and Evaluated Models}
\label{sec:generation:baselines}

Table~\ref{tab:models} shows the models we use for generation. 
We select one strong proprietary model (\texttt{Gemini Flash 2.5}) and two strong open-source models (\texttt{Qwen 3.5-9B} and \texttt{Gemma 4-E4B}) for answer generation (Prompt~\ref{prompt:answer-generation} in the appendix). We use temperature of 0.\footnote{We re-run some examples for \texttt{Qwen} with a temperature of 0.05 when it entered thinking loops with a temperature of 0.} Each generator is paired with the strongest retrievers of the same model family.

\subsection{Evaluation Metrics} 
\label{sec:generation:eval_metrics}

We test the generated answers along two dimensions: (a) how similar they are to ground truth answers, and (b) how grounded they are in retrieved documents. The latter measures the inverse of hallucination rate. We use a variation of FActScore \cite{min-etal-2023-factscore} together with \texttt{Gemini-2.5-Pro}.\footnote{To assess the robustness of the \texttt{Gemini} judge, we repeated the evaluation using \texttt{gpt-oss-120b} and \texttt{Qwen3.5-122B-A10B} as judges. We find that their claim-level judgements moderately to strongly correlate with \texttt{Gemini}'s at Pearson's $r=$ 0.68 and 0.73, respectively.} We decompose the generated answer $\hat{y}_q$ into a set of atomic claims $A(\hat{y}_q)=\{a_{q,1},\ldots,a_{q,m_q}\}$. We then look at the proportion of claims in $A(\hat{y}_q)$ that are supported by the evidence set, which is defined differently for each dimension below. 

\paragraph{Accuracy against ground truth answers.} The evidence set is $A(y_q)$, which is obtained by using the same LLM to decompose the gold standard answer $y_q$ into atomic claims. We then compute the accuracy of claims as follows: 
\begin{equation*}
\scalebox{0.85}{$ \displaystyle
Acc(\hat{y}_q, y_q)
 =
\frac{1}{|A(\hat{y}_q)|}
\sum_{a \in A(\hat{y}_q)}
\mathbb{I}
\left[
\max_{e \in A(y_q)}
\operatorname{Supp}(a,e)
=1
\right]
$}
\end{equation*}
\noindent where $\operatorname{Supp}(a,e)$ is an entailment-based support judgment returned by the factuality evaluator (see Prompts~\ref{prompt:factscore_groundedness}--\ref{prompt:factscore_entailment} in the appendix). A claim is counted as grounded if at least one claim in the evidence set $A(y_q)$ entails it. Following FActScore and X-FactAlign \cite{samir-etal-2024-locating}, this metric measures the fraction of atomic claims in the generated answer that are supported by the provided evidence, rather than independent legal correctness.

\paragraph{Groundedness in documents.} The evidence set is $E_q$, under condition $c \in \{\textsc{oracle}, \textsc{pipeline}\}$. In the pipeline condition, documents are gathered from retrieval methods: $E_q=D_q^{r}$. In the oracle condition, we confine retrieval to only chunks from ground truth documents: $E_q^{c}=D_q^{GT}$.\footnote{We used the Qwen Embedder 8192 chunk size with reranking configuration across oracle conditions for equal evaluation.} The pipeline condition measures the impact that retrieval has on final answers and partially measures the ability of retrieval methods despite missing documents in the gold set. Query-level groundedness is thus similarly defined as: 
\begin{equation*}
\scalebox{0.85}{$ \displaystyle
G(q,c)
= 
\frac{1}{|A(\hat{y}_q)|}
\sum_{a \in A(\hat{y}_q)}
\mathbb{I}
\left[
\max_{e \in E_q^{c}}
\operatorname{Supp}(a,e)
=1
\right]
$}
\end{equation*}

\begin{table}[t]
\centering
\footnotesize
\begin{tabular*}{\columnwidth}{@{\extracolsep{\fill}} llccc}
\toprule
\textbf{Cond.} & \textbf{LLM} & \textbf{$n$} & \textbf{Ground.} & \textbf{Acc.} \\
\midrule
Oracle & Gemini     & 532 & 0.7839      & 0.5806 \\
Oracle & Gemma      & 532 & \bf{0.7983} & \bf{0.6278} \\
Oracle & Qwen       & 526 & 0.7488      & 0.5790 \\
\midrule
Pipeline & Gemini   & 532 & \bf{0.7656} & 0.4329 \\
Pipeline & Gemma    & 532 & 0.7448      & \bf{0.4906} \\
Pipeline & Qwen     & 526 & 0.7103      & 0.4198 \\
\bottomrule
\end{tabular*}
\caption{Groundedness and accuracy scores across conditions. Bold numbers are best in the condition.}
\vspace{-5pt}
\label{tab:groundedness_eval}
\end{table}


\subsection{Results} 
\label{sec:generation:results}

\autoref{tab:groundedness_eval} presents groundedness and accuracy across generations and conditions. For each model family, we used the top configuration (Table~\ref{tab:main-results}). 

\paragraph{Models generate claims unsupported by documents.} Across all setups, at least 20\% of generated claims are unsupported by the documents.

In both conditions, \texttt{Qwen} creates the highest number of unsupported claims, possibly because its answers are longer on average (14,685 characters vs.\ 6,057 for \texttt{Gemma} and 11,550 for \texttt{Gemini}) and its reasoning can become fixated on incorrect ideas.

To investigate the nature of unsupported claims, we sampled 25 unsupported claims from the best-performing model in each condition: \texttt{Gemma} in the oracle condition and \texttt{Gemini} in the pipeline condition, and categorized them in Tables~\ref{tab:gemma_error_analysis} and \ref{tab:gemini_error_analysis} in Appendix, respectively. For \texttt{Gemma}, about 42\% are valid contextual or concluding statements absent from the retrieved documents, while 58\% are hallucinations, including confusion between general and case-specific law, misapplied legal tests, document misattribution, and errors involving details such as ``shall not'' vs.\ ``may not''. In the \texttt{Gemini-2} pipeline condition, around 60\% of the claims come from synthesizing texts and creating grounded statements too semantically far from the original texts. The other 40\% of the ungrounded claims were hallucinations. Extrapolating this to the results in Table~\ref{tab:groundedness_eval}, we estimate a lower bound of 8\% hallucination rate with an upper bound of 29\%.

\paragraph{Generated answers differ substantially from the gold standard, especially when based on different documents.} The accuracy of generated answers in the oracle condition is 58-63\%, demonstrating substantial deviation. We analyze unsupported claims in Appendix~\ref{sec:appendix:unsupported}. The accuracy is much lower in the pipeline condition, when it is based on different documents than the expert answer. \textbf{These findings illustrate the importance of a good retrieval component, as well as the need for reliable automatic metrics for legal hallucinations \cite{warranty}.}

\subsection{Analysis}
\label{sec:generation:analysis}

Table~\ref{tab:performance_analysis} shows that subgroup trends in generation partly follow retrieval nDCG trends. For example, queries simulating legal experts have higher nDCG@10 for all three family-matched pipelines, but this does not translate uniformly into better answers.

From layperson to expert queries, accuracy increases for \texttt{Gemini} and \texttt{Gemma} but decreases for \texttt{Qwen}. Groundedness increases for \texttt{Gemini} and \texttt{Qwen} but decreases for \texttt{Gemma}. For \texttt{Qwen}, nDCG@10 improves from 0.442 to 0.508 on expert queries while accuracy drops from 0.435 to 0.414, suggesting that legal phrasing improves retrieval, but not necessarily answer quality. However, automatic groundedness evaluation also contains some noise.

Jurisdictional results show a similar decoupling. \texttt{Gemini} has its highest nDCG@10 at 0.538 on the Other subset of provinces, yet performs worst there with 0.401 accuracy. For \texttt{Gemma}, groundedness and accuracy do not consistently correlate across jurisdictions. Groundedness decreases from Ontario to BC, increases in Alberta, and decreases again for Other, whereas accuracy increases from Ontario to BC and then decreases across Alberta and Other. \texttt{Qwen} has the most stable groundedness but the widest accuracy range: from 0.356 on Other to 0.461 on Ontario. All configurations vary widely across jurisdictions, highlighting the need for more robust, jurisdiction-specific systems.

\section{Conclusion}
\label{sec:conclusion}
We introduced \bench{}, a benchmark for evaluating RAG over Canadian case law. It differs from prior benchmarks by focusing on natural language queries simulating realistic common use of public-facing legal assistants. Evaluation on various RAG systems reveals high hallucination rates, but it also highlights the limitations of automatic evaluation of relevance and accuracy. Future work should develop additional benchmarks to cover other aspects of Canadian law, such as case laws in French and Quebec civil-law sources. By releasing the code and data, we aim to support transparent comparison and more reliable development of RAG systems for legal information tasks.

\section*{Limitations}
\paragraph{Data Coverage.} \bench{} only covers Canadian case law in select courts and provinces, and only focuses on English case laws. Future work can develop benchmarks that cover bilingual Canadian law, and Quebec civil-law sources, statute and regulation. In addition, due to the cost of annotations, our dataset is only meant for testing rather than training models.

\paragraph{Evaluation.} Our findings highlighted the challenge in evaluating document relevance and answer correctness automatically. To overcome these limitations, we employed both expert evaluation and various types of analyses. However, expert annotation is costly and doesn't scale. Future work should focus on developing reliable evaluation metrics for legal RAG. Moreover, our evaluation largely ignored the temporal relevance and the precedential value of courts, which could be important to determine document relevance. We leave this challenge to future work.

\section*{Ethics Statement}
\paragraph{Data Access.} We used two sources of case laws: the CLD dataset, which is publicly available, and a dataset which was provided to us by a private company with permission to publish it. The benchmark itself combines queries generated by an LLM with answers collected from experts, and we make it publicly available along with our code under an open MIT license. Data from courts are subject to their own licenses which have limitations for commercial use. By releasing the data, prompts, model outputs, and evaluation code, we aim to support transparent comparison and more reliable development of legal RAG systems for Canadian legal information tasks.

\paragraph{Data Privacy.} While case laws contain names and other personal information, all data we use in this work was already public record. 

\paragraph{Expert Annotation.} The work was approved by our institutional review board (IRB). We hired annotators at a university and through Upwork and compensated them \$20 CAD per hour for students and \$40 per hour for the professional paralegal. All rates are higher than the minimum wage. We explained to them the goal of improving automated systems for legal research and answer synthesis. To demonstrate how costly it is to collect a moderate size dataset like ours while still paying experts fairly, overall we spent around \$6,000 CAD on data collection and manual evaluation. We collected consent forms from annotators where we detailed the context of our research and publication goals. We highlighted the risk of viewing sensitive content in court cases, which they agreed to.

\paragraph{Implications.} Performance on \bench is not directly indicative of real-world performance, and we do not advocate for replacing legal expertise with AI. People should still consult lawyers when they face legal issues, and lawyers are still responsible for the quality and accuracy of their legal research, whether or not it was supported by AI tools. The current premise of AI-based legal research tools is that they save experts time. While we did not perform a controlled experiment, we expected the expert evaluation (\S\ref{sec:retrieval:results}) -- which is similar to a lawyer verifying AI-generated outputs -- to be faster (and cheaper) than our initial data collection (\S\ref{sec:data}) -- which simulates legal research. In practice, it took more time; verifying documents (from 3 methods) took experts over an hour per query on average. Our findings are in line with \newcite{choi2024lawyering} who showed that AI-based legal research tools may improve the quality of answers but do not increase productivity. In particular, we argue that because models make subtle errors, verification is a huge bottleneck in legal AI. Legal professionals relying on AI tools may incur ``verification debt'', where answer generation leads to short-term productivity gains which cancel out with manual and difficult verification \cite{hochberg2026ai}.

\section*{Acknowledgments}
We would like to thank the legal expert annotators: Hugo Mak, Hooria Hayat, Michelle Lee, Kiona Choi Smith for their contribution to our benchmark. This work was funded by NSERC Alliance and Mitacs Accelerate grants in collaboration with Caseway. The authors are also supported by the Vector Institute, Canada CIFAR AI Chairs program, and NSERC Discovery grants. This research was enabled in part by computational resources and services provided by the Digital Research Alliance of Canada, Gemini credits through Google's Gemini Academic Program Award, and Kanon 2 credits from Isaacus AI.

\bibliography{references_for_arxiv}

\appendix
\clearpage
\newpage
\onecolumn
\section{Annotation Details}

We accepted students who had above-average grades and/or prior work experience under a lawyer. Annotators were provided a UI that presented all the queries for a given seed query. They were instructed to first flag and skip nonsensical, ill-formed or very complex queries. Then, they were tasked with searching for relevant case laws. We allowed annotators to use common search tools that they would normally use in legal research, such as the Canadian Legal Information Institute (\href{https://www.canlii.org/}{CanLii}) and Thomson Reuters' \href{https://legal.thomsonreuters.com/en/westlaw}{Westlaw}.\footnote{While the search in these tools may involve AI, annotators were still responsible for constructing the queries and manually selecting among the retrieved cases. To improve the completeness of the final answers, we also allowed them to also cite laws, but we required that each query have at least one case law.}
For each case law they retrieved, they were asked to select relevant text of each document. After completing the retrieval stage, they moved on to the answer phase. To speed up the annotation process, instead of asking annotators to phrase an answer from scratch, we provided the query and the selected relevant texts to \texttt{Gemini 2.5 Pro} to generate an answer, and instructed the annotators to verify or edit to a final answer. This design was informed by consultation with credible legal experts, who indicated that searching for relevant case laws is the most challenging step of legal research. To prevent AI influence in the retrieval stage, once the annotators moved on to the answer generation stage, the retrieval stage was locked. 

The following are the original instructions given to annotators. We received feedback from the annotators and emailed clarifications and updates throughout. For example, part way through, we lifted the max document limit while stating they should still aim to include less than 10 documents per query on average. We also allowed them to use statutes part way through to improve the speed, though we still required at least 1 case law per query.

\begin{prompt}[title={Annotation Guidelines}, label=prompt:annotation_guidelines, listing options={basicstyle=\scriptsize\ttfamily, breaklines=true, breakindent=0pt}]
    Load a Batch: Enter a batch ID below. Each batch contains 8 unique legal queries. There are 2 fact variations per batch. Each has 4 situation variations which aim to keep the base facts the same.
    Track Your Time: Before starting each session, make sure to start a timer. We use this to track how long it takes to complete each batch so we can pay you appropriately!
    Evaluate the Query: Read the queries and evaluate them using the provided form. If a query is too vague or too complex, you can flag it and skip it. If it is obviously too complex to be completed within a reasonable time, you can skip it. If more than 2 queries are flagged, just skip the batch. After doing a few, you can get a sense of how long each one should take and adjust your approach accordingly.
    Gather Evidence: Read each query and use external resources (WestLaw, CanLII, etc.) to find relevant cases for the given jurisdiction. Only use court cases or regulations. Add your source documents to the query, including the case title, citation, and URL.
    
    Target Scope:
    Every 4 queries should take around 40-80 minutes (less is great). If it consistently takes longer, use fewer sources. Do not use more than 10 sources (pick the best ones). Do not include excessively long documents that take a long time to read if possible (use your judgement). We are targeting to have each student complete around 4-8 batches per week.
    Extract & Paste: Read your sources and copy the highly relevant quotes into the "Snippets" section (one point per snippet). Copy the whole contents of the original source and paste it into the "Whole Document Text" field. If there isn't a good way to get the text, just do `Ctrl+A` and adjust so that the main text is selected. Do not spend much time on copying or formatting the sources!
    Reuse Work: Use the Copy to... menu on a document to instantly duplicate it across multiple queries.
    AI Synthesis & Final Answer Standards: Use the Custom Instructions box to guide the AI's focus. Click either "Generate (Snippets Only)" or "Generate (Include Full Text)". If the full text is too long, the tool will warn you to use Snippets Only.
    
    Important: Document Locking
    Generating an AI answer will permanently lock your documents for that query to prevent biasing your research based on the AI's output. You can still read the documents, but you cannot edit, add, or delete them after generation. Make sure your research is complete and you are ready before generating!
    
    What makes a good Final Answer?
    Required structure - three sections:
    Opening Statements: Introduce the topic/area of law, paraphrase the question to make the legal issue clear, give a short hedge of the conclusion.
    Supporting Arguments: Arguments and evidence drawn from the gathered documents, with discussion of how they support or qualify the answer.
    Final Conclusion: A clear concluding statement synthesizing the above.
    Synthesize, don't just answer: Focus on what *previously happened* in court, rather than giving a definitive "yes" or "no".
    Acknowledge Missing Facts: If important query details are NOT covered by your evidence, explicitly state this and explain why those missing facts might matter. For example:
    "Based on the provided evidence, it appears that courts determine that factor X was the most important in reaching the outcome (2022 ONCA 45). However, there is no evidence about factor Y, which could also be important because..."
    "The most relevant pieces of evidence are (2022 ONCA 45) which says X and (2019 BCCA 12) which says Y. Based on your case, it's not clear which way it could go - some things you should keep in mind are..."
    Exact Citations: Citations must appear exactly as entered in the Citation field. If a citation cannot be verified quickly, use neutral citation format (e.g. Case Name, 2022 ONCA 45) or remove it.
    Stay Grounded: Rely ONLY on the provided evidence and very general legal knowledge. Omit introductory filler.
    Crucial Step: The AI is just a drafter. You must read, edit, and manually verify the generated answer to ensure it meets these standards and accurately satisfies the query.
\end{prompt}
\FloatBarrier
\newpage
\section{Prompt Templates}
\label{appendix:PromptTemplates}

\begin{prompt}[title={Prompt \thetcbcounter: Query Generation Example}, label=prompt:query_generation, listing options={basicstyle=\scriptsize\ttfamily}] 
Your input fields are:
1. 'court_decision_text' (str): A full Canadian legal court decision.
2. 'user_persona' (str): The type of user using a Legal AI tool and their intent (e.g., 'Layperson looking 
for the outcome', 'Lawyer looking for citations').
3. 'user_traits' (str): A description of the user's personality.
4. 'target_section' (str): The section of the document to focus on: Overview, Reasoning, Decision.

Your output fields are:
1. 'generated_query' (str): A realistic search query that the user might type into a Legal AI tool, 
based on their persona and the target section.

All interactions will be structured in the following way, with the appropriate values filled in.

[[ ## court_decision_text ## ]]
{court_decision_text}

[[ ## user_persona ## ]]
{user_persona}

[[ ## user_traits ## ]]
{user_traits}

[[ ## target_section ## ]]
{target_section}

[[ ## generated_query ## ]]
{generated_query}

[[ ## completed ## ]]

In adhering to this structure, your objective is: 
Given a court decision and a specific user persona/intent and target section of the document (Overview, 
Reasoning, Decision), generate a realistic search query that a user might type, and the corresponding 
factual answer found strictly within the document around the target section.

The generated query must be a realistic search query or question relevant to the text but assuming no 
knowledge of the document. The query should be asked from the point of view of a user doing exploratory 
research on their issue. Do NOT reference specific names, dates, entities, or other specific case details. 
Create queries that a user might ask *without* knowing which case document contains the answer. The 
queries should reflect the legal issues, principles, fact patterns, and situations discussed in the 
provided court decision, but MUST NOT reveal or directly quote the document.

The Query must have at least 2-3 unique facts or distinguishing details that are relevant to the document 
but do not directly reveal the document. For example, if the document is about a car accident case, 
the query could be "What happens if on a snowy day where the conditions were poor, I'm involved in a 
rear-end collision and the other driver was texting?"

Introduce variety in the types of queries generated, including but not limited to:
- Hypothetical situations ("If I ...", "What happens when ...?")
- Conceptual legal query (How does the court determine...?")
- Outcome-oriented query ("Would this count as...?")
- Multi-step or scenario-based query requiring reasoning.
- The query should be moderately complex, not trivial.

[[ ## court_decision_text ## ]]
<FULL COURT DECISION TEXT GOES HERE>

[[ ## user_persona ## ]]
Layperson: A non-legal expert asking a general query about a realistic legal scenario they may face in the 
real world. They do not use legal jargon.

[[ ## user_traits ## ]]
User traits: openness: low, conscientiousness: medium, extraversion: high, agreeableness: high, 
neuroticism: high. 

Query Style Instructions: Do NOT cite specific case names, statute numbers, or legal tests. Use perfect 
grammar, precise legal terminology, and complex sentence structures. Use standard, functional grammar. 
Use loose grammar, potential typos, simple keywords, or fragment sentences. Frame the query as a 'rambling' 
or 'venting' personal story. Include specific details of what went wrong to give context, rather than 
just asking the legal question directly. Be terse, brief, and to the point. Tone should be urgent, 
worried, seeking reassurance, or focusing on risks/penalties. Tone should be calm, detached, and 
objective. Focus on abstract concepts, reasoning, implications, or 'why' questions. Focus strictly on the
tangible 'Adverse Event' or 'Injury' described in the text (e.g., 'forced to retake class', 'denied 
refund', 'basement flooded'). Do not use abstract summaries like 'school problem' or 'career help'. 
Use polite, soft openers and tone (e.g., 'I'm hoping you can help...'), but ensure the core complaint 
describes the specific unfair event mentioned in the text. Phrasing should be demanding, skeptical, or 
aggressive (e.g., 'Prove that...').

[[ ## target_section ## ]]
Reasoning

Respond with the corresponding output fields, starting with the field  [[ ## generated_query ## ]], and 
then ending with the marker for  [[ ## completed ## ]].
\end{prompt}

\begin{prompt}[title={Prompt \thetcbcounter: Query Variations - Generate Alternative Facts}, label=prompt:query_variation_alternative_facts]
=== Message 0 | Role: system ===
Your input fields are:
1. `original_query` (str): The original query for which to generate a fact variation.
Your output fields are:
1. `reasoning` (str):
2. `query_variation` (str): The generated query variation based on the input original query with altered facts but the same scenario type and tone/style.
All interactions will be structured in the following way, with the appropriate values filled in.

[[ ## original_query ## ]]
{original_query}

[[ ## reasoning ## ]]
{reasoning}

[[ ## query_variation ## ]]
{query_variation}

[[ ## completed ## ]]
In adhering to this structure, your objective is:
Creating queries to evaluate a RAG tool that helps everyday people navigate legal issues.
Receives a query and generates a variation on the facts of the query while keeping the same scenario type and tone/style. The fact variation should be a realistic alteration of the original facts that still fits within the same general scenario type.
The query variation should change what court cases would be relevant to the query, but should not be so outlandish as to change the scenario type. The variations should also not change the tone and style of the query.
The query variation should not increase the complexity of the query or make it more difficult to understand. The query variation should be something that a person might realistically ask about in relation to the same type of legal issue.

Examples:
1. Original query: "My kid, he's 8, got into a fight and punched someone just one time. But the person got seriously hurt. Now we're dealing with court. Will he go to jail? Like, what kind of punishment could he get for causing bad injuries from one punch? I'm so worried.""
Fact variation example: ""My kid, he's 17, got into a fight and punched someone just one time. But the person got seriously hurt. Now we're dealing with court. Will he go to jail? Like, what kind of punishment could he get for causing bad injuries from one punch? I'm so worried.""

2. Original query: "My ex keeps making up stories that I hurt our child, even when all the investigations show nothing happened. It's so frustrating because they keep involving child services and now everyone thinks *her* constant talk about abuse is actually making our child suffer emotionally. What happens if one parent won't stop pushing false allegations and it causes harm to the child, and what can be done to protect the child from their other parent's actions like that?"
Fact variation example: "My ex keeps making up stories that I hurt our child. It's so frustrating because they keep involving child services and now everyone thinks *my* constant talk about abuse is actually making our child suffer emotionally. What happens if one parent won't stop pushing false allegations and it causes harm to the child, and what can be done to protect the child from their other parent's actions like that?"

=== Message 1 | Role: user ===
[[ ## original_query ## ]]
{query}Respond with the corresponding output fields, starting with the field `[[ ## reasoning ## ]]`, then `[[ ## query_variation ## ]]`, and then ending with the marker for `[[ ## completed ## ]]`.
\end{prompt}

\clearpage

\begin{prompt}[title={Prompt \thetcbcounter: Query Variations - Generate Alternative Situation}, label=prompt:query_variation_alternative_situation, listing options={basicstyle=\scriptsize\ttfamily}]
=== Message 0 | Role: system ===
Your input fields are:
1. `query` (str): The original query for which to generate variations.
2. `query_variation_type` (str): The type of variation to create
3. `variation_description` (str): A description of the type of variation to create
4. `query_variation_example` (str): An example of the type of variation to create
5. `original_query_example` (str): An example of an original query that the variation example is based on
6. `query_variation_tone_style` (str): The tone and style to use when creating the query variation. 
For example, 'layperson' or 'legal expert'.
Your output fields are:
1. `reasoning` (str):
2. `query_variation` (str): The generated query variation based on the input query, variation type, 
and tone/style.
All interactions will be structured in the following way, with the appropriate values filled in.

[[ ## query ## ]]
{query}

[[ ## query_variation_type ## ]]
{query_variation_type}

[[ ## variation_description ## ]]
{variation_description}

[[ ## query_variation_example ## ]]
{query_variation_example}

[[ ## original_query_example ## ]]
{original_query_example}

[[ ## query_variation_tone_style ## ]]
{query_variation_tone_style}

[[ ## reasoning ## ]]
{reasoning}

[[ ## query_variation ## ]]
{query_variation}

[[ ## completed ## ]]
In adhering to this structure, your objective is:
Creating queries to evaluate a RAG tool that helps everyday people navigate legal issues.
The queries will be annotated with legal experts and are thus costly to annotate. To reduce cost, 
the variations should require minimal changes to the court case documents that are required to answer the 
query. The variations should not be so outlandish as to create unrealistic queries that people would never 
actually ask about legal issues. The variations should also not change the tone and style of the query 
unless specified to be in the tone of a legal expert.

Receives a query and variation type and tone/style. Then creates a variation of the query.

=== Message 1 | Role: user ===
[[ ## query ## ]]
{query}

[[ ## query_variation_type ## ]]
{variation_type}

[[ ## variation_description ## ]]
{variation_description}

[[ ## query_variation_example ## ]]
{variation_example}

[[ ## original_query_example ## ]]
What are my rights if my landlord enters my apartment without notice?

[[ ## query_variation_tone_style ## ]]
{layperson/legal expert}

Respond with the corresponding output fields, starting with the field `[[ ## reasoning ## ]]`, 
then `[[ ## query_variation ## ]]`, and then ending with the marker for `[[ ## completed ## ]]`.
\end{prompt}

\begin{prompt}[title={Prompt \thetcbcounter: LLM Judge for Query Creation - System Prompt}, label=prompt:judge-system]
Your input fields are:
1. 'document_snippet' (str): The legal text serving as the ground truth answer.
2. 'query' (str): The generated user question to be evaluated.
3. 'criteria_guidelines' (str): The specific rules and examples this pair must satisfy.
Your output fields are:
1. 'reasoning' (str): 
2. 'analysis' (str): A step-by-step analysis of how the pair meets or fails EACH criterion.
3. 'final_verdict' (str): Must be exactly 'Keep' or 'Reject'.
All interactions will be structured in the following way, with the appropriate values filled in.

[[ ## document_snippet ## ]]
{document_snippet}

[[ ## query ## ]]
{query}

[[ ## criteria_guidelines ## ]]
{criteria_guidelines}

[[ ## reasoning ## ]]
{reasoning}

[[ ## analysis ## ]]
{analysis}

[[ ## final_verdict ## ]]
{final_verdict}

[[ ## completed ## ]]
In adhering to this structure, your objective is: 
You are an expert Legal Retrieval Judge. 
You will audit a (Query, Document) pair to decide if it belongs in a Gold Standard
Test Set for a realistic Legal Retrieval system that retrieves documents based on
questions people have about the law or legal precedents.
The QUERY is a user query that should be answerable by retrieving the DOCUMENT.
The DOCUMENT is a legal court decision that serves as the ground truth document to
retrieve in a Legal Retrieval system.

You must evaluate the pair against the provided CRITERIA GUIDELINES.
If the pair fails ANY of the criteria, the final verdict must be 'Reject'.
\end{prompt}

\clearpage

\begin{prompt}[title={Prompt \thetcbcounter: LLM Judge for Query Creation - User Prompt}, label=prompt:judge-user]
[[ ## document_snippet ## ]]
<the actual document text>

[[ ## query ## ]]
<the actual user query>

[[ ## criteria_guidelines ## ]]
### Structural Integrity (No AI Artifacts)
**Rule**: The query must not contain 'AI-isms', references to 'the provided text', or explicit request for summaries. It should sound like a human asking a question, not a prompt engineer.
**Examples**:
- [FAIL]: Summarize the provided legal document regarding the plaintiff.
- [FAIL]: Based on the text below, what was the ruling?
- [FAIL]: Extract the key dates from this case file.
- [PASS]: What is the precedent for constructive dismissal in Ontario?
- [PASS]: Can a landlord evict a tenant for personal use if they own another property?

### Information Gap (The 'Google Test')
**Rule**: The query must be specific enough to require a legal retrieval tool. If the
query is trivial generic knowledge (answerable by a quick Google search without this
doc), it fails.
**Examples**:
- [FAIL]: What does the Supreme Court do?
- [FAIL]: Define 'contract'.
- [FAIL]: Who is the current Prime Minister?
- [PASS]: Does the Oakes test apply to administrative tribunals?
- [PASS]: limitations period for medical malpractice in British Columbia 2023

### Relevance & Grounding
**Rule**: The document must actually contain the answer or be highly relevant precedent
for the query. The query must not hallucinate facts not present in the document.
**Examples**:
- [FAIL]: Query asks about a 'drunk truck driver' when the document is about a 'slip and fall' in a grocery store.
- [FAIL]: Query asks for the 2024 statute amendment, but the document is a 1990 case decision.
- [PASS]: Query asks about 'slip and fall liability' and the document is a ruling on 'Doe v. Walmart' regarding a wet floor.
- [PASS]: Query asks for 'grounds for patent invalidation' and the document outlines a decision on 'obviousness' in a patent suit.

### Tax court specific test for relevant cases
**Rule**: The document must be about self represented litigants, informal procedures, tax dispute resolution mechanisms or other disputes that are relevant to everyday
people. This is specifically about the type of query and not about matching the query to the document. If the document is about a tax court case that is very technical and not relevant to everyday people, it fails.
**Examples**:
- [FAIL]: The document is about a complex corporate tax dispute between two multinational corporations that has no relevance to everyday people.
- [FAIL]: The document is about a tax court case that is very technical and not relevant to everyday people.
- [PASS]: The document is about a tax court case involving a self represented litigant disputing a CRA assessment for a small business.
- [PASS]: The document is about a tax court case involving an individual disputing the disallowance of a home office deduction.

Respond with the corresponding output fields, starting with the field [[ ## reasoning ## ]], then [[ ## analysis ## ]], then [[ ## final_verdict ## ]], and then ending with the marker for [[ ## completed ## ]].
\end{prompt}

\clearpage

\begin{prompt}[title={Prompt \thetcbcounter: Iter-RetGen}, label=prompt:iter-retgen]
System:
You are a legal research assistant drafting an intermediate answer that will be used to refine the next retrieval round.
Answer the question accurately and concisely using ONLY the provided context passages. Focus on what courts have decided previously rather than issuing a definitive yes / no. If the passages do not contain enough information for some aspect of the question, say so clearly and name the missing aspect so the next search can target it. Cite the relevant passage(s) when possible.

User:
Context:
[1] ({citation})
{chunk_text}

[2] ({citation})
{chunk_text}

...

Question: {query_text}
\end{prompt}

\begin{prompt}[title={Prompt \thetcbcounter: Answer Generation}, label=prompt:answer-generation]
You are a legal research assistant answering questions about Canadian law 
using ONLY the provided context passages and very general legal knowledge.
Do not focus on giving a definitive 'yes' or 'no' answer. Synthesise the 
evidence into a clear, concise response that describes what courts have 
previously decided on similar issues.
If important details in the question are NOT covered by the passages, state 
explicitly that the evidence is insufficient for those aspects and explain 
why those missing facts could matter. Do not invent information to fill gaps.
Structure your answer in exactly three sections using these plain-text headings 
(no markdown):
1. Opening Statements
- Introduce the topic and general area of law.
- Paraphrase the question to make the legal issue clear.
- Give a short hedge of the conclusion.
2. Supporting Arguments
- Arguments and evidence drawn from the provided passages.
- Discussion of how the evidence supports or qualifies the answer.
3. Final Conclusion
- A clear concluding statement synthesising the above.
CITATION FORMAT: Cite sources using the exact citation string shown in each 
passage header (e.g. '2022 ONCA 45'). Do not include the case name, paragraph, 
section, or page references. Do not paraphrase or invent citations. If no 
citation is available, omit the reference.
Omit introductory filler.

Context:
{context}

Question: {query}
\end{prompt}

\begin{prompt}[title={Prompt \thetcbcounter: Response Groundedness}, label=prompt:factscore_groundedness]
You are a world class expert designed to evaluate the groundedness of an assertion.
You will be provided with an assertion and a context.
Your task is to determine if the assertion is supported by the context.
Follow the instructions below:
A. If there is no context or no assertion or context is empty or assertion is empty, say 0.
B. If the assertion is not supported by the context, say 0.
C. If the assertion is partially supported by the context, say 1.
D. If the assertion is fully supported by the context, say 2.
You must provide a rating of 0, 1, or 2, nothing else.
Return your response as JSON in this format: {"rating": X} where X is 0, 1, or 2.
Please return the output in a JSON format that complies with the following schema as specified in JSON Schema:
{"description": "Structured output for response groundedness evaluation.", "properties": {"rating": {"description": "Groundedness rating (0, 1, or 2)", "title": "Rating", "type": "integer"}}, "required": ["rating"], "title": "ResponseGroundednessOutput", "type": "object"}Do not use single quotes in your response but double quotes,properly escaped with a backslash.

--------EXAMPLES-----------
Example 1
Input: {
    "response": "Albert Einstein was born in Germany.",
    "context": "Albert Einstein was born March 14, 1879 at Ulm, in Wurttemberg, Germany."
}
Output: {
    "rating": 2
}

Example 2
Input: {
    "response": "Einstein was a chemist who invented gunpowder.",
    "context": "Albert Einstein was a theoretical physicist known for his theory of relativity."
}
Output: {
    "rating": 0
}

Example 3
Input: {
    "response": "Einstein received the Nobel Prize.",
    "context": "Albert Einstein received the 1921 Nobel Prize in Physics for his services to theoretical physics."
}
Output: {
    "rating": 2
}
-----------------------------

Now perform the same with the following input
input: {
    "response": "<GENERATED_ANSWER>",
    "context": "<GOLD_ANSWER>"
}
Output:
\end{prompt}

\begin{prompt}[title={Prompt \thetcbcounter: Response Groundedness Judge}, label=prompt:factscore_judge]
As a specialist in assessing the strength of connections between statements and their given contexts, I will evaluate the level of support an assertion receives from the provided context. Follow these guidelines:

* If the assertion is not supported or context is empty or assertion is empty, assign a score of 0.
* If the assertion is partially supported, assign a score of 1.
* If the assertion is fully supported, assign a score of 2.

I will provide a rating of 0, 1, or 2, without any additional information.
Return your response as JSON in this format: {"rating": X} where X is 0, 1, or 2.
Please return the output in a JSON format that complies with the following schema as specified in JSON Schema:
{"description": "Structured output for response groundedness evaluation.", "properties": {"rating": {"description": "Groundedness rating (0, 1, or 2)", "title": "Rating", "type": "integer"}}, "required": ["rating"], "title": "ResponseGroundednessOutput", "type": "object"}Do not use single quotes in your response but double quotes,properly escaped with a backslash.

--------EXAMPLES-----------
Example 1
Input: {
    "response": "Albert Einstein was a scientist.",
    "context": "Albert Einstein was a German-born theoretical physicist widely held to be one of the greatest and most influential scientists of all time."
}
Output: {
    "rating": 2
}

Example 2
Input: {
    "response": "Einstein invented television.",
    "context": "Albert Einstein developed the theory of relativity."
}
Output: {
    "rating": 0
}

Example 3
Input: {
    "response": "Einstein won a Nobel Prize.",
    "context": "Albert Einstein received the 1921 Nobel Prize in Physics."
}
Output: {
    "rating": 2
}
-----------------------------

Now perform the same with the following input
input: {
    "response": "<GENERATED_ANSWER>",
    "context": "<GOLD_ANSWER>"
}
Output:
\end{prompt}

\begin{prompt}[title={Prompt \thetcbcounter: Claim Decomposition}, label=prompt:factscore_decompose]
Decompose and break down each of the input sentences into one or more standalone statements. Each statement should be a standalone claim that can be independently verified.
Follow the level of atomicity and coverage as shown in the examples.
Please return the output in a JSON format that complies with the following schema as specified in JSON Schema:
{"description": "Output from claim decomposition.", "properties": {"claims": {"description": "Decomposed claims", "items": {"type": "string"}, "title": "Claims", "type": "array"}}, "required": ["claims"], "title": "ClaimDecompositionOutput", "type": "object"}Do not use single quotes in your response but double quotes,properly escaped with a backslash.

--------EXAMPLES-----------
Example 1
Input: {
    "response": "Charles Babbage was a French mathematician, philosopher, and food critic.",
    "atomicity": "low",
    "coverage": "low"
}
Output: {
    "claims": [
        "Charles Babbage was a mathematician and philosopher."
    ]
}

Example 2
Input: {
    "response": "Albert Einstein was a German theoretical physicist. He developed the theory of relativity and also contributed to the development of quantum mechanics.",
    "atomicity": "low",
    "coverage": "low"
}
Output: {
    "claims": [
        "Albert Einstein was a German physicist.",
        "Albert Einstein developed relativity and contributed to quantum mechanics."
    ]
}
-----------------------------

Now perform the same with the following input
input: {
    "response": "<GENERATED_ANSWER>",
    "atomicity": "low",
    "coverage": "low"
}
Output:
\end{prompt}

\begin{prompt}[title={Prompt \thetcbcounter: Entailment-based Support Judgement}, label=prompt:factscore_entailment, listing options={basicstyle=\scriptsize\ttfamily}]
Your task is to judge the faithfulness of a series of statements based on a given context. For each 
statement you must return verdict as 1 if the statement can be directly inferred based on the context or 0 
if the statement can not be directly inferred based on the context.
Please return the output in a JSON format that complies with the following schema as specified in JSON 
Schema:
{"\$defs": {"StatementFaithfulnessAnswer": {"description": "Individual statement with reason and verdict 
for NLI evaluation.", "properties": {"statement": {"description": "the original statement, word-by-word", 
"title": "Statement", "type": "string"}, "reason": {"description": "the reason of the verdict", "title": 
"Reason", "type": "string"}, "verdict": {"description": "the verdict(0/1) of the faithfulness.", "title": 
"Verdict", "type": "integer"}}, "required": ["statement", "reason", "verdict"], "title": 
"StatementFaithfulnessAnswer", "type": "object"}}, "description": "Structured output for NLI statement 
evaluation.", "properties": {"statements": {"items": {"\$ref": "#/\$defs/StatementFaithfulnessAnswer"}, 
"title": "Statements", "type": "array"}}, "required": ["statements"], "title": "NLIStatementOutput", 
"type": "object"}

Do not use single quotes in your response but double quotes, properly escaped with a backslash.

--------EXAMPLES-----------
Example 1
Input: {
    "context": "John is a student at XYZ University. He is pursuing a degree in Computer Science. He is 
    enrolled in several courses this semester, including Data Structures, Algorithms, and Database 
    Management. John is a diligent student and spends a significant amount of time studying and completing 
    assignments. He often stays late in the library to work on his projects.",
    "statements": [
        "John is majoring in Biology.",
        "John is taking a course on Artificial Intelligence.",
        "John is a dedicated student.",
        "John has a part-time job."
    ]
}
Output: {
    "statements": [
        {
            "statement": "John is majoring in Biology.",
            "reason": "John's major is explicitly stated as Computer Science, not Biology.",
            "verdict": 0
        },
        {
            "statement": "John is taking a course on Artificial Intelligence.",
            "reason": "The context mentions courses in Data Structures, Algorithms, and Database 
            Management, but does not mention Artificial Intelligence.",
            "verdict": 0
        },
        {
            "statement": "John is a dedicated student.",
            "reason": "The context states that John is a diligent student who spends a significant amount 
            of time studying and completing assignments.",
            "verdict": 1
        },
        {
            "statement": "John has a part-time job.",
            "reason": "There is no information in the context about John having a part-time job.",
            "verdict": 0
        }
    ]
}
-----------------------------

Now perform the same with the following input
input: {
    "context": "<CONTEXT>",
    "statements": [
        "<CLAIM_1>",
        "<CLAIM_2>",
        "...",
        "<CLAIM_N>"
    ]
}
Output:
\end{prompt}
\newpage
\section{Additional Information}
\label{appendix:additional_info}
\FloatBarrier

\subsection{Full Dataset Statistics}
\label{appendix:full_dataset_stats}
\begin{table}[t]
    \centering
    \footnotesize
    \begin{tabular}{lr}
        \toprule
        \textbf{Statistic} & \textbf{Value} \\
        \midrule
        Total queries & 532 \\
        Total ground truth (query, document) pairs & 3{,}193 \\
        Total unique ground truth documents & 588 \\
        \midrule
        Mean ground truth docs per query & 6.02 \\
        Std.\ dev.\ ground truth docs per query & 4.47 \\
        Min.\ ground truth docs per query & 1 \\
        Median ground truth docs per query & 4 \\
        Max.\ ground truth docs per query & 23 \\
        \midrule
        \textbf{Jurisdictions} \\ 
        \midrule
        Ontario & 28\% \\
        British Columbia & 22\% \\
        Alberta & 22\% \\
        Other & 28\% \\
        \bottomrule
    \end{tabular}
    \vspace{-5pt}
    \caption{Data statistics of \bench{}.}
    \label{tab:stats}
\end{table}
\FloatBarrier

\subsection{Court Statistics}
\label{appendix:additional_info:court}

\begin{table}[h]
    \centering
    \small
    \begin{tabular}{lrlrlr}
        \toprule
        \textbf{Court} & \textbf{Proportion} & \textbf{Court} & \textbf{Proportion} & \textbf{Court} & \textbf{Proportion}\\
        \midrule
        BCSC & 11.55\% &
        ONSC & 10.12\% &
        SCC & 9.49\% \\
        UNKNOWN & 9.31\% &
        TCC & 6.80\% &
        STATUTE & 4.66\% \\
        CHRT & 4.66\% &
        SST & 4.48\% &
        ABQB & 3.40\% \\
        FCA & 3.31\% &
        BCCA & 3.04\% &
        FCT & 2.33\% \\
        ONCJ & 2.24\% &
        ABPC & 1.88\% &
        ABCA & 1.79\% \\
        HRTO & 1.61\% &
        FC & 1.52\% &
        BCPC & 1.52\% \\
        SKCA & 1.52\% &
        ONCA & 1.25\% &
        NSSC & 1.25\% \\
        SKQB & 1.07\% &
        AHRC & 0.90\% &
        NWTSC & 0.81\% \\
        NBQB & 0.72\% &
        NLSC & 0.63\% &
        NLTD & 0.63\% \\
        ABKB & 0.54\% &
        QCCQ & 0.54\% &
        NLCA & 0.45\% \\
        NLSCTD & 0.45\% &
        NBKB & 0.45\% &
        QCCS & 0.45\% \\
        ONSCDC & 0.45\% &
        QCCA & 0.36\% &
        NBCA & 0.36\% \\
        NSCA & 0.36\% &
        PESCTD & 0.36\% &
        SKKB & 0.36\% \\
        NLPC & 0.36\% &
        NSPC & 0.27\% &
        YKCA & 0.18\% \\
        BCHRT & 0.18\% &
        SKPC & 0.18\% &
        ALRB & 0.09\% \\
        YKSC & 0.09\% &
        NWTCA & 0.09\% &
        PECA & 0.09\% \\
        PESCAD & 0.09\% &
        NWTTC & 0.09\% &
        BCCRT & 0.09\% \\
        NSSM & 0.09\% &
        MBCA & 0.09\% &
        YKTC & 0.09\% \\
        MBQB & 0.09\% &
        ABCJ & 0.09\% &
        ONSCSM & 0.09\% \\
        CM & 0.09\% \\
        \bottomrule
    \end{tabular}
    \caption{Proportion of courts in the retrieval base. Courts from human annotators were parsed from the citation however, some were not recoverable due to conflicting formats.}
    \label{tab:court_stats}
\end{table}
\FloatBarrier

\subsection{Province Statistics}
\label{appendix:additional_info:provinces}

\begin{table}[h]
    \centering
    \small
    \begin{tabular}{lr}
        \toprule
        \textbf{Province} & \textbf{Proportion} \\
        \midrule
        Ontario & 28.01\% \\
        British Columbia & 22.37\% \\
        Alberta & 21.62\% \\
        Saskatchewan & 8.46\% \\
        Northwest Territories & 5.45\% \\
        Newfoundland and Labrador & 4.32\% \\
        Prince Edward Island & 3.01\% \\
        Nova Scotia & 2.63\% \\
        New Brunswick & 2.26\% \\
        Yukon & 1.88\% \\
        
        \bottomrule
    \end{tabular}
    \caption{Proportion of queries from each province.}
    \label{tab:province_stats}
\end{table}
\newpage
\subsection{Retrieval Corpus Statistics}
\label{appendix:full_corpus_stats}
\begin{table}[h]
    \centering
    \small
    \begin{tabular}{ll}
    \toprule
    \textbf{Metric} & \textbf{Value} \\
    \midrule
    Federal                    & 38.0\% \\
    British Columbia           & 19.0\% \\
    Ontario                    & 18.3\% \\
    Alberta                    & 10.1\% \\
    Saskatchewan               & 3.6\% \\
    Newfoundland and Labrador  & 2.9\% \\
    Nova Scotia                & 2.3\% \\
    New Brunswick              & 1.8\% \\
    Quebec                     & 1.6\% \\
    Northwest Territories      & 1.1\% \\
    Prince Edward Island       & 0.6\% \\
    Yukon                      & 0.4\% \\
    Manitoba                   & 0.2\% \\
    
    \midrule
    Total documents  & 1{,}117 \\
    Year range       & 1940--2026 \\
    Median Year      & 2014 \\
    Average Words      & 7{,}550 \\
    Total words      & $\approx$ 8.4M \\
    \bottomrule
  \end{tabular}
  \caption{Retrieval Corpus Statistics.}
  \label{tab:corpus-stats}
\end{table}

\subsection{Query Variations}
\label{appendix:additional_info:variations}

\begin{table}[ht]
\centering
\small
\begin{tabular}{L{3cm}L{5cm}L{6cm}}
\toprule
\textbf{Variation Type} & \textbf{Description} &  \textbf{Example} \\
\midrule
Change Point of View & Changes perspective of the asker & As a landlord, what can happen if I enter my tenant's apartment without notice? \\
Next Steps Focus & Refocuses on practical procedure, action steps or finding help & What can I do if my landlord enters my apartment without notice? \\
Rule Focus & Targets legal rules/principles at play & What determines if a landlord needs to give notice before entering a tenant's apartment \\
Information Seeking & Expressing request for information/understanding the scenario & Do courts generally require landlords to give notice when entering a tenant's apartment? \\
Burden of Proof & Focusing on what evidence is needed in the given scenario & What kind of evidence would I need to show entry without notice? \\
General Interpretations of a Legal Principle & Generalizes from the specific situation to broader interpretations & How do courts interpret the requirement for landlords to give notice before entering their tenant's apartment? \\
\bottomrule
\end{tabular}
\caption{All query variations used to vary seed queries in the query generation phase, using the example seed query ``What are my rights if my landlord enters my apartment without notice?''.}
\label{tab:query_variations}
\end{table}
\FloatBarrier
\clearpage
\newpage
\section{Models Used}
\begin{table}[h]
\centering
\small
\setlength{\tabcolsep}{2pt} 
\begin{tabular}{lll}
\toprule
\textbf{Model} & \textbf{Type} & \textbf{Task} \\
\midrule
\texttt{Qwen3-Embedding-8B} & Open & Embedding \\
\texttt{EmbeddingGemma} & Open & Embedding \\
\texttt{Gemini-Embedding-001} & Closed & Embedding \\
\texttt{Gemini-Embedding-2} & Closed & Embedding \\
\texttt{Kanon-2-Embedding} & Closed & Embedding \\
\texttt{Kanon-2-Reranker} & Closed & Reranking \\
\midrule
\texttt{Qwen-3.5-9B} & Open & Text generation \\
\texttt{Gemma-4-E4B} & Open & Text generation \\
\texttt{Gemini-Flash-2.5} & Closed & Text generation \\
\texttt{Gemini-Pro-2.5} & Closed & \makecell[l]{Text generation,\\ Judging}\\
\bottomrule
\end{tabular}
\caption{Evaluated models and categories. All open embedding models run on A6000s for 0.5-2 hours. Open Generator models ran for 8-24 hours.}
\label{tab:models}
\end{table}
\section{Additional Results}
\label{appendix:additional_results}

\subsection{Full Results}
\label{appendix:full_results}
\begin{table*}[t]
\centering
\renewcommand{\arraystretch}{0.9}
\resizebox{\textwidth}{!}{%
\begin{tabular}{lclcrrrrr}
\toprule
\textbf{LLM} & \textbf{Chunk Size} & \textbf{Method} & \textbf{With Rerank} & \textbf{MRR} & \textbf{Recall@10} & \textbf{Recall@25} & \textbf{nDCG@10} & \textbf{nDCG@25} \\
\midrule
\multicolumn{9}{l}{\textit{\textbf{BM25}}} \\
\quad BM25 & DocLevel & -- & \textcolor{RedOrange}{\XSolidBrush} & 0.245 & 0.193$^*$ & 0.309$^*$ & 0.163$^*$ & 0.208$^*$ \\
\quad BM25 & 1024 & -- & \textcolor{RedOrange}{\XSolidBrush} & 0.197 & 0.113 & 0.195 & 0.119 & 0.147 \\
\quad BM25 & 4096 & -- & \textcolor{RedOrange}{\XSolidBrush} & 0.256$^*$ & 0.152 & 0.251 & 0.162 & 0.197 \\
\midrule
\multicolumn{9}{l}{\textit{\textbf{Hybrid}}} \\
\quad Qwen & 1024 & Hybrid BM25 & \textcolor{RedOrange}{\XSolidBrush} & 0.483 & 0.299 & 0.454 & 0.331 & 0.375 \\
\quad Qwen & 4096 & Hybrid BM25 & \textcolor{RedOrange}{\XSolidBrush} & 0.516 & 0.320 & 0.487 & 0.360 & 0.405 \\
\quad Qwen & 1024 & Hybrid BM25 + Kanon2Rerank & \textcolor{ForestGreen}{\Checkmark} & 0.563$^*$ & 0.364$^*$ & 0.548$^*$ & 0.426$^*$ & 0.477$^*$ \\
\midrule
\multicolumn{9}{l}{\textit{\textbf{Qwen}}} \\
\quad Qwen & 1024 & -- & \textcolor{RedOrange}{\XSolidBrush} & 0.561 & 0.359 & 0.520 & 0.419 & 0.463 \\
\quad Qwen & 2048 & -- & \textcolor{RedOrange}{\XSolidBrush} & 0.594 & 0.354 & 0.524 & 0.429 & 0.474 \\
\quad Qwen & 4096 & -- & \textcolor{RedOrange}{\XSolidBrush} & 0.577 & 0.367 & 0.555 & 0.433 & 0.480 \\
\quad Qwen & 8192 & -- & \textcolor{RedOrange}{\XSolidBrush} & 0.594 & 0.393 & 0.582 & 0.446 & 0.495 \\
\quad Qwen & 1024 & -- & \textcolor{ForestGreen}{\Checkmark} & 0.563 & 0.363 & 0.548 & 0.426 & 0.477 \\
\quad Qwen & 2048 & -- & \textcolor{ForestGreen}{\Checkmark} & 0.584 & 0.373 & 0.571 & 0.445 & 0.497 \\
\quad Qwen & 4096 & -- & \textcolor{ForestGreen}{\Checkmark} & 0.581 & 0.374 & 0.588 & 0.439 & 0.497 \\
\quad Qwen & 8192 & -- & \textcolor{ForestGreen}{\Checkmark} & 0.594 & 0.417$^*$ & \underline{0.636}$^*$ & 0.456 & 0.522 \\
\quad Qwen & 4096 & IterRetGen & \textcolor{RedOrange}{\XSolidBrush} & 0.616 & 0.387 & 0.577 & 0.461 & 0.509 \\
\quad Qwen & 4096 & IterRetGen & \textcolor{ForestGreen}{\Checkmark} & 0.613 & 0.377 & 0.577 & 0.455 & 0.509 \\
\quad Qwen & 8192 & IterRetGen & \textcolor{ForestGreen}{\Checkmark} & 0.632$^*$ & 0.406 & 0.601 & 0.474$^*$ & 0.527$^*$ \\
\midrule
\multicolumn{9}{l}{\textit{\textbf{Gemma}}} \\
\quad Gemma & 1024 & -- & \textcolor{RedOrange}{\XSolidBrush} & 0.602 & 0.349 & 0.521 & 0.422 & 0.464 \\
\quad Gemma & 2048 & -- & \textcolor{RedOrange}{\XSolidBrush} & 0.570 & 0.367 & 0.537 & 0.417 & 0.461 \\
\quad Gemma & 4096 & -- & \textcolor{RedOrange}{\XSolidBrush} & 0.583 & 0.365 & 0.529 & 0.419 & 0.458 \\
\quad Gemma & 8192 & -- & \textcolor{RedOrange}{\XSolidBrush} & 0.574 & 0.393 & 0.567 & 0.426 & 0.472 \\
\quad Gemma & 4096 & -- & \textcolor{ForestGreen}{\Checkmark} & 0.584 & 0.384 & 0.598 & 0.443 & 0.502 \\
\quad Gemma & 8192 & -- & \textcolor{ForestGreen}{\Checkmark} & 0.598 & 0.419 & \textbf{0.643}$^*$ & 0.458 & 0.527$^*$ \\
\quad Gemma & 4096 & IterRetGen & \textcolor{RedOrange}{\XSolidBrush} & 0.615 & 0.393 & 0.553 & 0.457 & 0.488 \\
\quad Gemma & 8192 & IterRetGen & \textcolor{RedOrange}{\XSolidBrush} & 0.624$^*$ & 0.429 & 0.595 & 0.473$^*$ & 0.512 \\
\quad Gemma & 4096 & IterRetGen & \textcolor{ForestGreen}{\Checkmark} & 0.619 & 0.391 & 0.551 & 0.457 & 0.489 \\
\quad Gemma & 8192 & IterRetGen & \textcolor{ForestGreen}{\Checkmark} & 0.613 & \underline{0.431}$^*$ & 0.595 & 0.471 & 0.510 \\
\midrule
\multicolumn{9}{l}{\textit{\textbf{Gemini1}}} \\
\quad Gemini1 & 1024 & -- & \textcolor{RedOrange}{\XSolidBrush} & 0.601 & 0.343 & 0.505 & 0.441 & 0.475 \\
\quad Gemini1 & 4096 & -- & \textcolor{RedOrange}{\XSolidBrush} & 0.625 & 0.378 & 0.557 & 0.472 & 0.507 \\
\quad Gemini1 & 8192 & -- & \textcolor{RedOrange}{\XSolidBrush} & 0.639 & 0.413 & 0.599$^*$ & 0.487 & 0.526 \\
\quad Gemini1 & 1024 & -- & \textcolor{ForestGreen}{\Checkmark} & 0.577 & 0.372 & 0.553 & 0.440 & 0.490 \\
\quad Gemini1 & 4096 & IterRetGen & \textcolor{RedOrange}{\XSolidBrush} & \underline{0.639} & 0.389 & 0.568 & 0.489 & 0.526 \\
\quad Gemini1 & 8192 & IterRetGen & \textcolor{RedOrange}{\XSolidBrush} & \textbf{0.655}$^*$ & 0.429$^*$ & 0.597 & \textbf{0.508}$^*$ & \underline{0.538}$^*$ \\
\midrule
\multicolumn{9}{l}{\textit{\textbf{Gemini2}}} \\
\quad Gemini2 & 4096 & -- & \textcolor{RedOrange}{\XSolidBrush} & 0.625 & \textbf{0.449} & 0.626 & \underline{0.489} & 0.534 \\
\quad Gemini2 & 8192 & -- & \textcolor{RedOrange}{\XSolidBrush} & 0.617 & 0.373 & 0.521 & 0.396 & 0.438 \\
\quad Gemini2 & 4096 & -- & \textcolor{ForestGreen}{\Checkmark} & 0.600 & 0.422 & \underline{\textbf{0.652}}$^*$ & 0.470 & \textbf{0.538} \\
\quad Gemini2 & 8192 & -- & \textcolor{ForestGreen}{\Checkmark} & 0.596 & 0.423 & 0.564 & 0.434 & 0.467 \\
\quad Gemini2 & 4096 & IterRetGen & \textcolor{RedOrange}{\XSolidBrush} & \underline{\textbf{0.661}}$^*$ & \underline{\textbf{0.456}}$^*$ & 0.631 & \underline{\textbf{0.521}}$^*$ & \underline{\textbf{0.560}}$^*$ \\
\midrule
\multicolumn{9}{l}{\textit{\textbf{Kanon2}}} \\
\quad Kanon2 & 1024 & -- & \textcolor{RedOrange}{\XSolidBrush} & 0.536 & 0.329 & 0.500 & 0.385 & 0.431 \\
\quad Kanon2 & 4096 & -- & \textcolor{RedOrange}{\XSolidBrush} & 0.546 & 0.336 & 0.541 & 0.394 & 0.453 \\
\quad Kanon2 & 8192 & -- & \textcolor{RedOrange}{\XSolidBrush} & 0.521 & 0.352 & 0.560 & 0.388 & 0.451 \\
\quad Kanon2 & 1024 & -- & \textcolor{ForestGreen}{\Checkmark} & 0.559 & 0.365 & 0.547 & 0.424 & 0.473 \\
\quad Kanon2 & 4096 & -- & \textcolor{ForestGreen}{\Checkmark} & 0.573 & 0.364 & 0.572 & 0.429 & 0.486 \\
\quad Kanon2 & 8192 & -- & \textcolor{ForestGreen}{\Checkmark} & 0.590$^*$ & 0.406$^*$ & 0.624$^*$ & 0.449$^*$ & 0.514$^*$ \\
\quad Kanon2 & 4096 & IterRetGen & \textcolor{RedOrange}{\XSolidBrush} & 0.577 & 0.369 & 0.555 & 0.431 & 0.479 \\
\quad Kanon2 & 8192 & IterRetGen & \textcolor{RedOrange}{\XSolidBrush} & 0.564 & 0.394 & 0.577 & 0.439 & 0.483 \\
\bottomrule
\end{tabular}}
\caption{Retrieval metrics across all experiments. Higher is better for all metrics. Bold Underlined = overall best, Bold = 2nd, Underlined = 3rd. $^*$ denotes best within group.}
\label{tab:retrieval_results}
\end{table*}
\FloatBarrier

\begin{figure}[ht]
    \centering
    \includegraphics[width=1.0\linewidth]{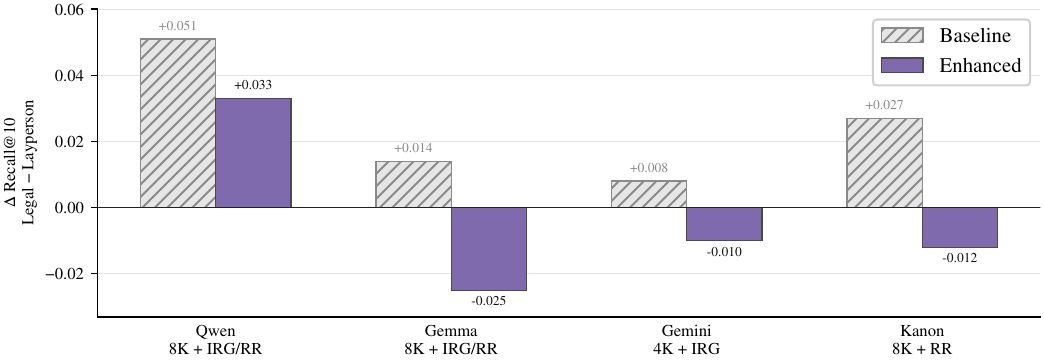}
    \caption{The average difference between Recall@10 of retrieval methods on legal and layperson style queries. The selected configurations are the base dense retrieval and the high performing configurations of the different embedding models. Positive values indicate a the retrieval system had higher performance on legal style queries.}
    \label{fig:style_recall}
\end{figure}

\begin{figure}[ht]
    \centering
    \includegraphics[width=1.0\linewidth]{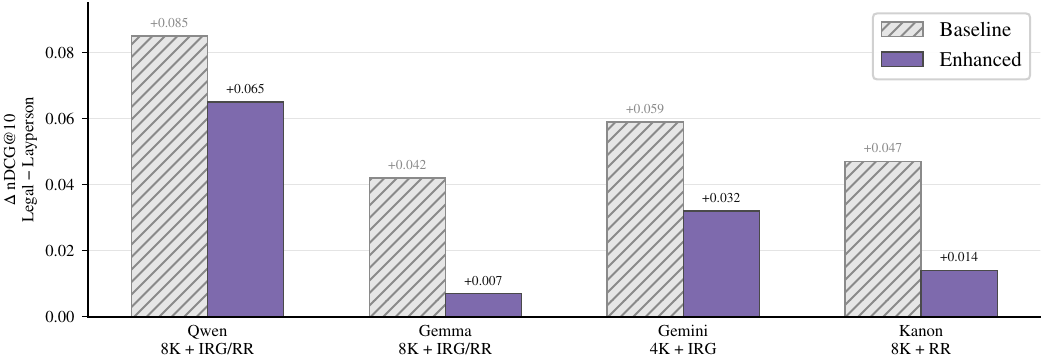}
    \caption{The difference between nDCG@10 of retrieval methods on legal and layperson style queries. The selected configurations are the base dense retrieval and the high performing configurations of the different embedding models. Positive values indicate a the retrieval system had higher performance on legal style queries.}
    \label{fig:style_ndcg}
\end{figure}

\subsection{Categorization of Unsupported Claims (Accuracy)}
\label{sec:appendix:unsupported}

\begin{table*}[]
\centering
\small
\begin{tabular}{lrl}
\toprule
\textbf{Error Category} & \textbf{Percentage} & \textbf{Description} \\
\midrule
\textbf{Too detailed} & 24.0\% & Contained more details that were not necessary  \\

\textbf{Irrelevant} & 24.0\% & Discussed irrelevant facts about other issues \\

\textbf{Misses the point} & 32.0\% & Answered other questions not helpful to the query  \\

\textbf{Relevant} & 4.0\% & Actually relevant but missing from the gold answer  \\

\textbf{Hallucination} & 8.0\% & The statement was not in the answer, nor was it grounded in retrieved documents.\\

\textbf{False Positive} & 8.0\% & The statement was judged as not being in the gold answer but was actually \\
\bottomrule
\end{tabular}
\caption{Categorization and distribution of 25 \texttt{Qwen} (oracle) generated statements which were judged as not being entailed by the statements in the ground truth answer.}
\label{tab:qwen_error_analysis}
\end{table*}

Of the three models, \texttt{Qwen} achieves the lowest accuracy. To better understand the type of errors that \texttt{Qwen} made, we sampled 25 unsupported claims and manually analyzed them (\autoref{tab:qwen_error_analysis}). We found that \texttt{Qwen}'s answers were significantly longer than the gold standard answers (that had an average of 3883.5 characters), and often contained claims that were either irrelevant, too detailed, or relevant but miss the point of the question (such as talking about factors that impacted other applications for employment insurance when it's already established the client is ineligible). A small number of errors include hallucinations (i.e. claims unsupported by the documents they cite), false positives (claims that appeared in the gold answer), and relevant claims missing from the gold answer. 

\newpage
\subsection{Categorization of Unsupported Claims (Groundedness)}
We can get very rough hallucination percentage by multiplying the 'true' hallucination ratio in the sample of 25 by the un-groundedness rate. This gives us around 8\% as a lower bound.
\label{sec:appendix:hallucination_categorization}

\begin{table*}[t]
\centering
\small
\renewcommand{\arraystretch}{1.5}
\setlength{\tabcolsep}{4pt}
\begin{tabularx}{\linewidth}{p{0.18\linewidth}c>{\hsize=0.92\hsize\raggedright\arraybackslash}X>{\hsize=1.08\hsize\raggedright\arraybackslash}X}
\toprule
\textbf{Error Category} & \textbf{Percentage} & \textbf{Description} & \textbf{Example} \\
\midrule
\textbf{Conversational Penalty} & 38.46\% & The LLM judge penalizes the generator for discussing opening or synthesizing documents in the concluding the answer. & An answer contained ``The inquiry concerns the employment insurance eligibility of a part-time employee who voluntarily left after a reduction in hours'' which was not a statement in the oracle documents. Another generated claim mentioned having to do a case-by-case basis due to different documents having different outcomes based on different facts. This `case-by-case' synthesis was judged as ungrounded\\
\midrule
\textbf{Judge Decontextualization Error} & 3.85\% & The decontextualizer confused entities in the claim & The government was swapped for the employer \\
\midrule
\textbf{Hierarchical Flattening} & 11.54\% & The generated answer elevates non-binding text, minority opinions, or specific factual reasoning to authoritative statutory law. & A generated answer confused the minority opinion for the majority opinion binding law in Canada v. Vavilov \\
\midrule
\textbf{Misapplication of Law} & 11.54\% & The generated answer contains misapplications of the law. & Applies incorrect test for litigation privilege to determine solicitor-client privilege because they both appear in the retrieved document \\
\midrule
\textbf{Scope Shifts} & 7.69\% & The generated answer softens a strict law or turns an option into a definite prerequisite. & A retrieved document says `shall not' definitively but the generated answer says `may not'. \\
\midrule
\textbf{False Insufficient Evidence} & 7.69\% & The generated answer incorrectly claims the retrieved context lacks the answer. & A generated answer correctly finds the burden of proof in the reasoning (supporting arguments) but in the conclusion states that there is not enough evidence to determine the burden of proof. \\
\midrule
\textbf{Entity Confusion} & 7.69\% & The generated answer confuses entities. & A generated answer said the applicant has a specific responsibility but the retrieved document explicitly says the Commission (government) has that responsibility. \\
\midrule
\textbf{Cites Wrong Document} & 7.69\% & The generated answer states a grounded fact but misattributes it to a different retrieved document which is not related to that fact. & In a generated statement discussing the value of the context of tweets, it cites a document which never discusses tweets or Social Media, while a separate retrieved document extensively discusses the topic. \\
\midrule
\textbf{Confused about Document Formatting} & 3.85\% & The generated answer misattributes the section of a rule due to formatting artifacts in the retrieved chunk. & The section number of a rule appeared in the document header while the section occurred partway through the page, the section number from the header was placed before the actual section. \\
\bottomrule
\end{tabularx}
\caption{Categorization and distribution of 25 Gemma generation statements which were judged as ungrounded by the oracle retrieved documents. Around 40\% are over penalizations while the rest illustrate a variety of hallucinations that LLMs face in the legal domain. Note that one claim was categorized into two error types.}
\label{tab:gemma_error_analysis}
\end{table*}

\begin{table*}[t]
\centering
\small
\renewcommand{\arraystretch}{1.5}
\begin{tabularx}{\linewidth}{p{0.18\linewidth}c>{\hsize=0.92\hsize\raggedright\arraybackslash}X>{\hsize=1.08\hsize\raggedright\arraybackslash}X}
\toprule
\textbf{Error Category} & \textbf{Percentage} & \textbf{Description} & \textbf{Example} \\
\midrule
\textbf{Conversational Penalty} & 56.00\% & The LLM judge penalizes the generator for discussing opening or synthesizing documents in the concluding the answer. & A generated claim states that a significant reduction in hours is considered when assessing just cause, but the LLM judge flags it as ungrounded because ``reduction in hours'' is absent from the retrieved document which discusses conditions that significantly modify wages. \\
\midrule
\textbf{Other Judge Error} & 4.00\% & Other LLM judge errors & In one generated answer, one document discussed facts less related to the statement while another document contradicted it using more relevant facts but the LLM judge discussed that the statement was not grounded because the former document contradicted it. \\
\midrule
\textbf{Misapplication of Law} & 12.00\% & The generated answer contains misapplications of the law. & In one generated answer, a very long retrieved document placed at the start of the context explicitly stated that solicitor-client privilege prevents a certain disclosure, however the generated statement cites it and makes the opposite claim, accidentally negating it. \\
\midrule
\textbf{Misattributing Jurisdiction} & 4.00\% & The generated answer cites a document and mistakes its jurisdiction. & In one generated answer, it explicitly cites a BC supreme court case where the court and province is provided and attributes it to Yukon courts to tailor it to the query. \\
\midrule
\textbf{Scope Shifts} & 4.00\% & The generated answer softens a strict law or hardens a permissive statutory option into a mandatory prerequisite. & An example answer inaccurately swaps from ``must do so if requested'' to ``must do so'' \\
\midrule
\textbf{Cites Wrong Document} & 8.00\% & The generated answer states a grounded fact but misattributes it to a different retrieved document or an incorrect level of the legislative hierarchy. & An example answer cites the Alberta Human Rights Code instead of Act. \\
\midrule
\textbf{Creates assumption in Query} & 4.00\% & Assumes an important fact based on query statements that do not actually support the claim & In a generated answer, it assumes that the user suffered so much burnout as to cause mental disability despite just saying they took sick leave due to burnout without additional details. \\ 
\midrule
\textbf{Using External Knowledge} & 8.00\% & The generated answer uses knowledge from pretraining to supplement answer. & In one generated answer, the retrieved document stated ``law grants the specific power to prosecute actions on behalf of an estate to the estate's formal `executor or administrator''', the answer used external logic of ``an affirmative grant of authority to a specific legal role inherently excludes those who do not hold that role.'' to expand its point. \\
\bottomrule
\end{tabularx}
\caption{Categorization and distribution of 25 generated statements for the Gemini pipeline on \bench. Around 60\% of flagged errors are LLM judge over-penalizations due to conversational text or connecting text between documents, while the remaining errors capture genuine structural slips like doctrinal conflations and statutory misattributions.}
\label{tab:gemini_error_analysis}
\end{table*}

\clearpage

\end{document}